\documentclass[10pt,twocolumn,letterpaper]{article}

\usepackage[pagenumbers]{cvpr} 

\usepackage[dvipsnames]{xcolor}

\definecolor{cvprblue}{rgb}{0.21,0.49,0.74}
\usepackage[pagebackref,breaklinks,colorlinks,citecolor=cvprblue]{hyperref}

\def\paperID{144} 
\def\confName{3DV\xspace}
\def\confYear{2025\xspace}

\usepackage{makecell}
\usepackage{multirow}

\usepackage{algorithm}
\usepackage{algorithmic}

\usepackage{pifont}
\newcommand{\cmark}{\ding{51}}%
\newcommand{\xmark}{\ding{55}}%

\DeclareMathOperator*{\argmin}{arg\,min~}

\newcommand{\gspose}{GS-Pose\xspace}
\newcommand{\gom}{3DGO\xspace}

\newcommand{\boldparagraph}[1]{
\paragraph{\textbf{#1}}
\vspace{-0.6em}
}

\title{\gspose: Generalizable Segmentation-based 6D Object Pose Estimation \\ with 3D Gaussian Splatting}

\author{
Dingding Cai \\
Tampere University, Finland \\
{\tt\small dingding.cai@tuni.fi}
\and
Janne Heikkilä \\
University of Oulu, Finland \\
{\tt\small janne.heikkila@oulu.fi}
\and
Esa Rahtu \\
Tampere University, Finland \\
{\tt\small esa.rahtu@tuni.fi}
}

\begin{document}

\maketitle
\begin{abstract}
This paper introduces \gspose, a unified framework for localizing and estimating the 6D pose of novel objects. \gspose begins with a set of posed RGB images of a previously unseen object and builds three distinct representations stored in a database. At inference, \gspose operates sequentially by locating the object in the input image, estimating its initial 6D pose using a retrieval approach, and refining the pose with a render-and-compare method. The key insight is the application of the appropriate object representation at each stage of the process. In particular, for the refinement step, we leverage 3D Gaussian splatting, a novel differentiable rendering technique that offers high rendering speed and relatively low optimization time. Off-the-shelf toolchains and commodity hardware, such as mobile phones, can be used to capture new objects to be added to the database. Extensive evaluations on the LINEMOD and OnePose-LowTexture datasets demonstrate excellent performance, establishing the new state-of-the-art. 
Project page: \href{https://dingdingcai.github.io/gs-pose}{https://dingdingcai.github.io/gs-pose}.

\end{abstract}    
\section{Introduction}
\label{sec:intro}

Acquiring the 3D orientation and 3D location of an object based on RGB images is a long-standing and important problem in computer vision and robotics. This 6D pose information is vital in applications that interact with the physical world, such as robotic manipulation \cite{collet2011moped,deng2020self} and augmented reality \cite{marchand2015pose,su2019deep}. Popular pose estimation approaches are based on training instance-specific models, and they often assume the availability of an external object detector for detecting the object from input RGB images. While some works have proposed approaches to circumvent this problem \cite{shugurov2022osop,ornek2023foundpose,lin2023sam}, they often rely on high-fidelity 3D CAD models of the object, which can be expensive and time-consuming to acquire. 

Ideally, a new object should be learned from a casually captured set of RGB reference images without requiring any expensive model parameter optimization. Recently, Liu \etal \cite{liu2022gen6d} introduced a method called Gen6D in this direction. Gen6D works by extracting 2D feature maps from the reference images, which are subsequently utilized for various sub-tasks, including object localization, initial pose estimation, and pose refinement. However, relying only on 2D representation often leads to sub-optimal performance. Alternatively, OnePose \cite{sun2022onepose} and OnePose++ \cite{he2022onepose++} explicitly reconstruct a 3D point cloud from the reference images via local feature matching. The 6D pose is obtained using 2D-3D correspondence matching between the test image and the reference point cloud. The practical challenge is to obtain an accurate 3D point cloud representation, particularly for texture-less and symmetric objects. Furthermore, both approaches still rely on an external object detector for cropping out the object of interest, limiting their applicability in real-world scenarios.

\begin{figure*}[tb]
    \centering
     \includegraphics[width=.8\linewidth]{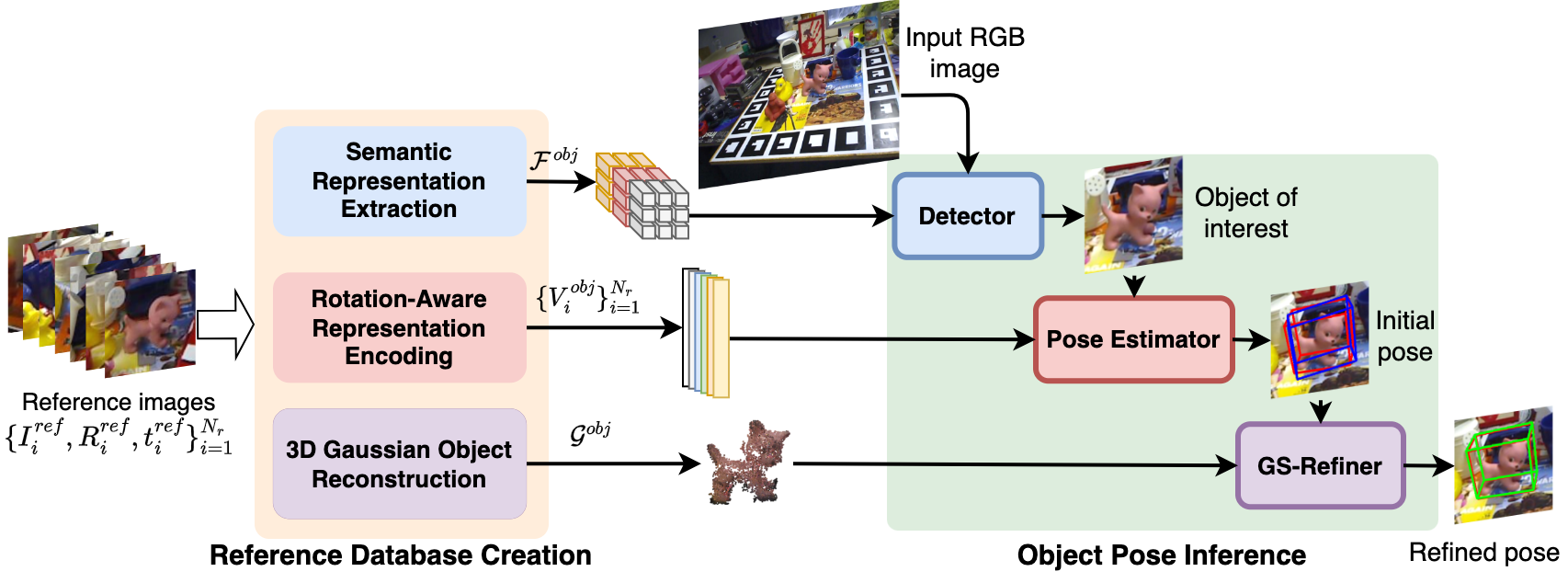}
    \caption{
    \textbf{Overview of \gspose}. \gspose involves two distinct phases to achieve pose estimation for a novel object, \ie, reference database creation and object pose inference. The first phase operates offline and occurs only once per object to construct multiple representations of the object. These representations include an object semantic representation ($\mathcal{F}^{obj}$), a set of rotation-aware embedding vectors ($\{V_i^{obj}\}^{N_r}_{i=1}$), and a 3D Gaussian Object ($\mathcal{G}^{obj}$). During inference, \gspose first employs an object detector to detect the object in a query image using the semantic information $\mathcal{F}^{obj}$. Then, \gspose adopts a pose estimator to produce an initial pose (blue box) from the detection result with the rotation-aware embeddings $\{V_i^{obj}\}^{N_r}_{i=1}$, Finally, \gspose leverages a pose refinement module (GS-Refiner) with $\mathcal{G}^{obj}$ to obtain a refined pose (green box). We indicate the ground-truth pose in red.
    }
    \label{fig-intro}
\end{figure*}

The key ingredient in 6D pose estimation is the object representation generated from the input images. Popular choices include 2D feature maps \cite{liu2022gen6d}, 3D point clouds\cite{sun2022onepose,he2022onepose++}, latent 3D models \cite{park2019latentfusion}, and 3D CAD models\cite{shugurov2022osop}, to name a few. Generally, each representation exhibits strengths in one aspect, \eg, object localization or fast initial 6D pose approximation, but performs poorly on other parts of the pipeline. With these insights, we propose a framework that applies multiple representations optimized for the three key steps: 1) object localization, 2) fast initial 6D pose estimation, and 3) iterative pose refinement. In particular, we leverage the recent advancements in so-called Foundation models and co-segmentation paradigms to construct powerful representations for object localization using only a handful of reference images. Secondly, we estimate a rough 6D pose using optimized template retrieval. Finally, the pose estimate is refined using an iterative render-and-compare technique. To this end, we rely on a novel inverse rendering method called 3D Gaussian Splatting (3DGS) \cite{kerbl20233d}, which represents a scene by many differentiable 3D Gaussian primitives with optimizable geometric and appearance properties. This explicit representation enables real-time photorealistic rendering capabilities, ideal for 6D pose optimization. 

We evaluate the proposed framework, called \gspose, on the LINEMOD\cite{hinterstoisser2012model} and OnePose-LowTexture\cite{he2022onepose++} datasets and obtain new state-of-the-art results on both benchmarks. 
The contributions of our work are summarized as follows:
\begin{itemize}
\item We present an integrated framework for 3D CAD model-free 6D object pose estimation. For each stage, we propose an optimized representation obtained from a set of posed RGB images of newly added objects. 
\item We present a generalizable co-segmentation approach for extracting object segmentation masks jointly from the reference RGB images, facilitating representation learning.
\item We present a robust 3D Gaussian splatting-based method for 6D object pose refinement. 
\item We experimentally confirm that the proposed framework achieves state-of-the-art performance on the LINEMOD and OnePose-LowTexture datasets. 
\end{itemize}

\section{Related Works}
\label{sec:review}


\noindent{\textbf{Object-Specific Pose Estimation.}}
Most existing pose estimation methods \cite{cai2022sc6d,chen2022epro,xiang2018posecnn,labbe2020cosypose,hodan2020epos,su2022zebrapose,gdrnet2021,li2019cdpn,haugaard2021surfemb,cai2023msda,peng2019pvnet} are object-specific pose estimators, which are specialized for pre-defined objects and cannot generalize to previously unseen objects without retraining. Some of them\cite{cai2022sc6d,xiang2018posecnn,labbe2020cosypose,gdrnet2021,cai2023msda,chen2022epro} directly regress the 6D pose parameters from RGB images by training deep neural networks on a large number of labeled images. While other approaches\cite{su2022zebrapose,haugaard2021surfemb,li2019cdpn,pix2pose2019,peng2019pvnet,hodan2020epos,chen2022epro} establish 2D-3D correspondences between 2D images and 3D object models to estimate the 6D pose by solving the Perspective-n-Point (PnP)\cite{lepetit2009epnp} problem. To relax the assumptions about each object instance, category-level methods\cite{wang2019normalized,chen2020learning,chen2020category,chen2021fs} have recently been proposed to handle unseen object instances of the same trained category by assuming that objects within the same category share similar shape priors. However, they are still incapable of estimating the object pose of unknown categories. 

\boldparagraph{Generalizable Object Pose Estimation.}
This type of work\cite{cai2022ove6d,he2022onepose++,liu2022gen6d,pan2023learning,shugurov2022osop,sun2022onepose,aae2018} removes the requirement of the object specific-training and can perform pose estimation for previously unseen objects during inference. There are two mainstreams, \ie, object model-based and object model-free. The model-based approaches \cite{cai2022ove6d,shugurov2022osop,aae2018} assume access to the 3D CAD models for rendering the object pose-conditioned images that are often utilized for template matching \cite{cai2022ove6d,aae2018,xiao2019pose}, pose refinement \cite{li2018deepim}, or correspondence establishment \cite{shugurov2022osop}. To avoid 3D CAD models, recent works \cite{he2022onepose++,sun2022onepose,liu2022gen6d,pan2023learning} resort to capturing object multi-view images with known poses as reference data for pose estimation. OnePose series \cite{he2022onepose++,sun2022onepose} utilize the posed RGB images to reconstruct 3D object point clouds and establish explicit 2D-3D correspondences between 2D query images and the reconstructed 3D point clouds to solve the 6D pose. However, reliance on correspondences becomes fragile when applied to objects with visual ambiguities, such as symmetry.
Besides, the above methods often assume that the 2D object detection or segmentation mask is available given a query image. 
In contrast, Gen6D \cite{liu2022gen6d} leverages the labeled reference images to detect the object in query images, initialize its pose, and then construct a 3D feature volume for pose refinement, which is the first work to simultaneously satisfy the requirements of being fully generalizable, model-free, and RGB-only. The follow-up works \cite{zhao2022finer,pan2023learning} revisit the Gen6D pipeline and improve the performance and robustness in object localization and pose estimation. 

\boldparagraph{2D Object Detection.} 
Commonly used object detection methods \cite{he2017mask,ren2015faster,redmon2016you} are category-specific detectors and cannot generalize to untrained categories. To tackle this issue, some approaches \cite{osokin2020os2d,zhao2022semantic,li2018high,liu2022gen6d,shugurov2022osop} leverage object reference images to detect previously unseen objects through template matching or feature correlation. However, they often show limited generalizability to new domains. 

\boldparagraph{3D Object Representation.} 
Most generalizable pose estimators \cite{cai2022ove6d,shugurov2022osop,ornek2023foundpose,lin2023sam,nguyen2022templates} often assume that the 3D object representations are available, such as 3D CAD models. OnePose family \cite{he2022onepose++,sun2022onepose} explicitly reconstructs 3D object point clouds from object multi-view RGB images, which can easily fail with challenging symmetric or textureless objects. Moreover, LatentFusion \cite{park2019latentfusion} and Gen6D series \cite{liu2022gen6d,pan2023learning} utilize the 2D image features to build the 3D object feature volumes for pose refinement. 
In this work, we instead exploit the differentiable 3D Gaussian Splatting \cite{kerbl20233d} technique to create 3D Gaussian Object representations for pose estimation. To the best of our knowledge, \gspose is the first work that leverages 3D Gaussian splatting for 6D object pose estimation.
    

\section{Approach}
\label{sec-method}

This section presents \gspose for estimating the 6D pose of novel objects from RGB images. An overview of \gspose is provided in \cref{fig-intro}.
\gspose operates in two distinct phases: object reference database creation and object pose inference. The creation phase, requiring RGB images of a novel object with known poses (\eg, captured with commodity devices like mobile phones), is performed offline once per object. During inference, \gspose leverages the pre-built object reference database to facilitate the 6D pose estimation task in a cascaded manner. 
In the subsequent subsections, 
we first present the reference database creation process in \cref{sec_MR}. 
Next, we describe the pose inference workflow in \cref{sec_infer}.
Finally, we present the objective functions for training \gspose in \cref{sec_loss}.

\subsection{Reference Database Creation}
\label{sec_MR}
\begin{figure*}[t]
    \centering
     \includegraphics[width=.8\linewidth]{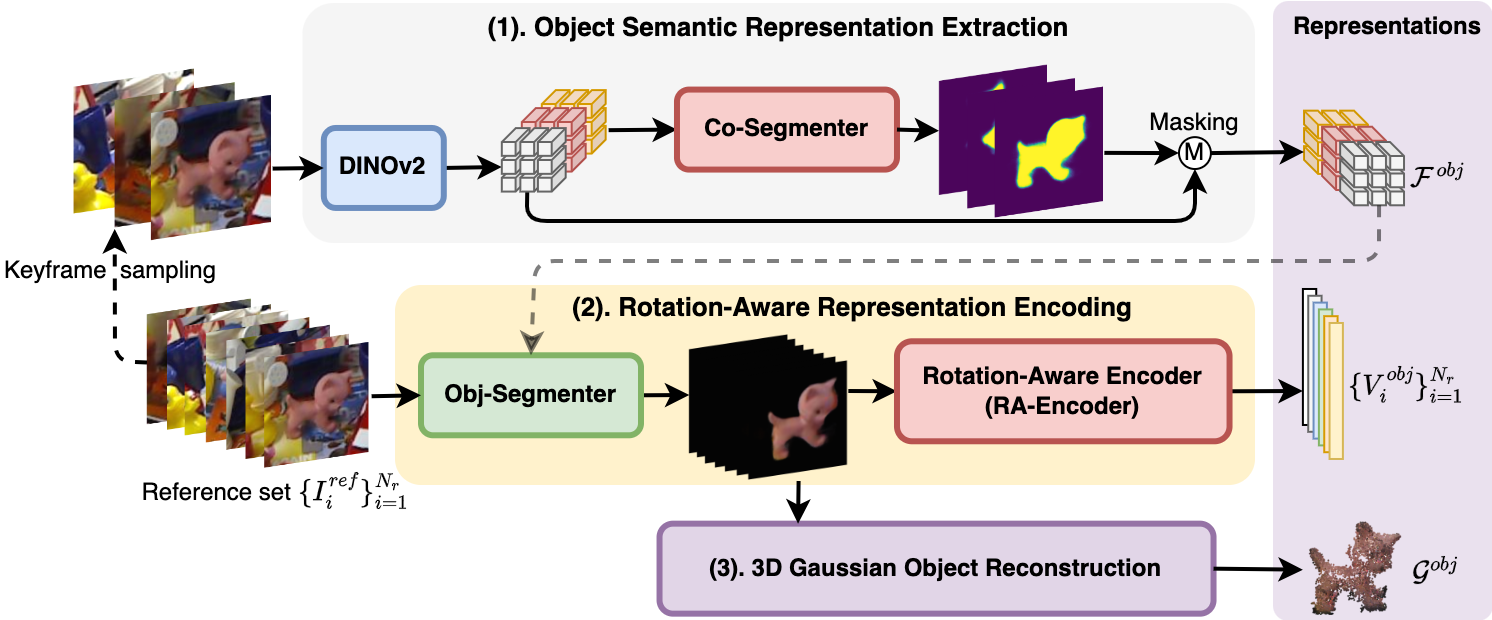}
    \caption{
    \textbf{Overview of the reference database creation process.} 
    We begin by selecting a group of keyframes from reference images.
    (1). These keyframes are processed through DINOv2 and Co-Segmenter to jointly predict object segmentation masks, which are then utilized to extract the object semantic tokens ($\mathcal{F}^{obj}$) from the keyframe features. 
    (2). Image-wise object segmentation is performed for all reference images $\{I^{ref}_i\}^{N_r}_{i=1}$ using an Obj-Segmenter with the obtained semantic information $\mathcal{F}^{obj}$. We then employ an RA-Encoder to extract the rotation-aware embeddings $\{V_i^{obj}\}^{N_r}_{i=1}$ from the segmented images.
    (3). Finally, we create a 3D Gaussian Object representation $\mathcal{G}^{obj}$ (viewed as a 3D point cloud for simplicity) using all segmented images with the known poses. 
    }
    \label{fig_creation}
\end{figure*}

This section describes the process for creating the reference database of a novel object based on its reference data.
This database is primarily comprised of object semantic representation $\mathcal{F}^{obj}$, a set of 3D object rotation-aware embedding vectors $\{V_i^{obj}\}^{N_r}_{i=1}$, and a 3D Gaussian Object representation $\mathcal{G}^{obj}$, where $N_r$ is the number of reference examples. The creation process involves three sub-steps: 
(1) semantic representation extraction, (2) 3D object rotation-aware representation encoding, and (3) 3D Gaussian Object (\gom) model reconstruction, as depicted in \cref{fig_creation}. 
In the following paragraphs, we elaborate on each sub-step.

\begin{figure}[tb]
    \centering
     \includegraphics[width=\linewidth]{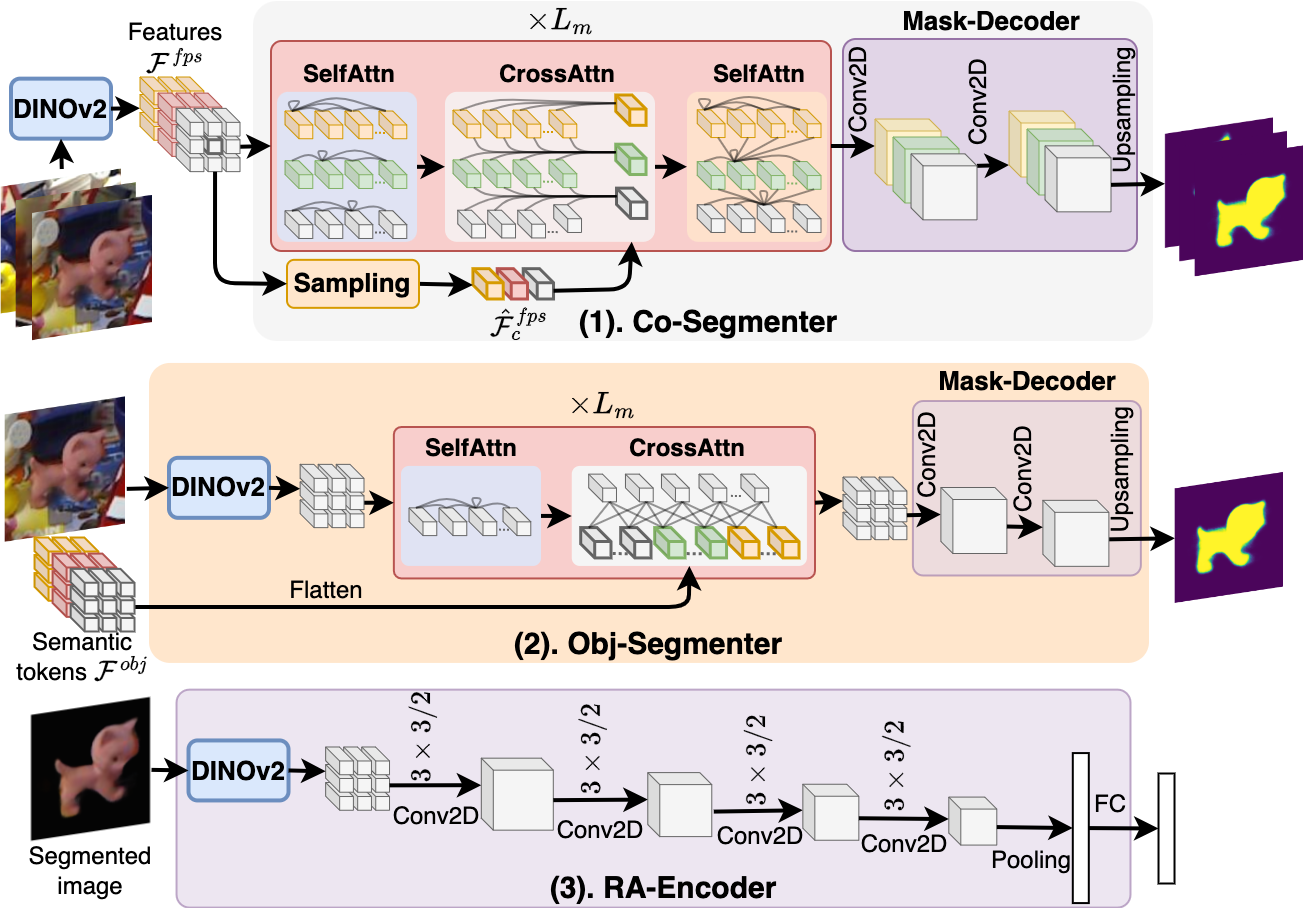}
    \caption{
    \textbf{(1). Co-Segmenter} includes a transformer-like module and a mask decoder to produce the co-segmentation masks. 
    \textbf{(2). Obj-Segmenter} consists of the DINOv2 backbone, a transformer-like module, and a mask decoder to predict the object mask.
    \textbf{(3). RA-Encoder} contains the DINOv2 backbone, four $3\times 3$ 2D convolutional (Conv2D) layers with stride 2, a generalizable average pooling layer, and a fully connected (FC) layer.
    }
    \label{fig_coseg}
\end{figure}

\boldparagraph{Semantic Representation Extraction.} 
To enable \gspose for 2D object detection and segmentation, we first extract a set of feature representation tokens that can effectively capture the semantic information of the target object from reference images. We leverage DINOv2\cite{oquab2023dinov2} to extract these tokens from RGB images. 
Essentially, a Co-Segmenter is employed to segment the object from the background, ensuring that only relevant feature tokens within the object region are considered (see \cref{fig_creation} top). 
Given $N_r$ reference images, we first select $N_k$ ($\ll N_r$) keyframes using farthest point sampling (FPS) \cite{qi2017pointnet++} based on their corresponding 3D rotation labels. 
Then, we extract image feature tokens $\mathcal{F}^{fps}\in \mathbb{R}^{N_k\times L\times C}$ from these keyframes using DINOv2, where $L$ and $C$ denote the token number and feature dimension of each frame. Next, we feed these feature tokens into the proposed Co-Segmenter, consisting of a transformer-like module and a mask decoding head, to jointly predict the object segmentation masks. 
Specifically, we reshape the keyframe feature tokens as feature maps 
(denoted as $\mathcal{\hat{F}}^{fps}$), 
from which we sample a set of frame-wise center tokens $\mathcal{\hat{F}}^{fps}_{c} \in \mathbb{R}^{N_k\times C}$ located at 
the 2D center of these feature maps. 
Next, the transformer-like module takes $\mathcal{F}^{fps}$ and $\mathcal{\hat{F}}^{fps}_c$ as input and sequentially performs \(L_m\) stacked self- and cross-attention computations (see \cref{fig_coseg} top).
The process can be formulated as
\begin{equation}
\small
\begin{split}
& L_m \times
\begin{cases}
    \mathcal{F}^{fps} = \text{SelfAttn}(\mathcal\mathcal{F}^{fps})            \in \mathbb{R}^{N_k\times L\times C} \\
    \mathcal{F}^{fps} = \text{Reshape}(\mathcal{F}^{fps})                    \in \mathbb{R}^{1\times N_kL\times C}        \\
    \mathcal{F}^{fps} = \text{CrossAttn}(\mathcal{F}^{fps}, \mathcal{\hat{F}}^{fps}_{c})                                   \\
    \mathcal{F}^{fps} = \text{SelfAttn}(\mathcal{F}^{fps})                   \in \mathbb{R}^{1\times 
N_kL\times C}        \\
    \mathcal{F}^{fps} = \text{Reshape}(\mathcal{F}^{fps})                    \in \mathbb{R}^{N_k\times L\times C} \\
\end{cases},
\end{split}
\label{eq:refMaskDec}
\end{equation}
where $L_m$ is the depth of the module.
The transformed $\mathcal{F}^{fps}$ is then fed into the mask decoding head (two $3\times 3$ convolutional layers followed by an upsampling layer) to produce the keyframe segmentation masks. 
Finally, we extract the object-aware semantic feature tokens $\mathcal{F}^{obj}$ from the keyframe feature maps $\mathcal{\hat{F}}^{fps}$ using the predicted masks.

\boldparagraph{Rotation-Aware Representation Encoding.} 
\label{sec-RA-encoder}
This step focuses on extracting the 3D object rotation-aware embedding vectors from reference images, which enables \gspose to estimate an initial pose via template retrieval. 
To achieve this, we first adopt an Obj-Segmenter to segment the object from each reference image and then utilize a Rotation-Aware Encoder (RA-Encoder) to extract an image-level embedding vector from the segmented image (see \cref{fig_creation} middle). 
Obj-Segmenter includes the DINOv2 backbone, a transformer-like module, and a mask decoding head (identical to the one in Co-Segmenter). 
Concretely, Obj-Segmenter first extracts the DINOv2 feature tokens $F^{ref}_i \in \mathbb{R}^{L\times C}$ from the $i^{th}$ reference image. 
Then, the image feature tokens ($F^{ref}_i$) along with the object semantic tokens ($\mathcal{F}^{obj}$) are fed into the transformer-like module to perform \(L_m\) stacked self- and cross-attention computations (see \cref{fig_coseg} middle).
This process can be formulated as
\begin{equation}
\small
\begin{split}
& L_m \times
\begin{cases}
    F^{ref}_i = \text{SelfAttn}(F^{ref}_i) \\ 
    F^{ref}_i = \text{CrossAttn}(F^{ref}_i, \mathcal{F}^{obj}) \\
\end{cases}.
\end{split} 
\label{eq:queMskDec}
\end{equation}
Subsequently, the mask decoding head is utilized to produce a segmentation mask $M^{ref}_i$ from the transformed image features $F^{ref}_i$.
Finally, we extract an image-level representation vector $V^{ref}_i \in \mathbb{R}^{64}$ from the segmented image using RA-Encoder.
RA-Encoder includes the DINOv2 backbone, four $3 \times 3$ convolutional layers with stride 2, a generalized average pooling layer, and a fully connected layer with an output dimension of 64 (see \cref{fig_coseg} bottom).

\boldparagraph{3D Gaussian Object Reconstruction.} 
The last step is to create the 3DGO representation $\mathcal{G}^{obj}$ for pose refinement (see \cref{fig_creation} bottom). 
3D Gaussian Splatting \cite{kerbl20233d} represents a 3D structure as a set of 3D Gaussians. Each 3D Gaussian is parameterized with a 3D coordinate $\mu \in \mathbb{R}^3$, a  3D rotation quaternion $r \in \mathbb{R}^4$, a scale vector $s \in \mathbb{R}^3$, an opacity factor $\alpha \in \mathbb{R}$, and spherical harmonics coefficients $h\in\mathbb{R}^k$, where $k$ is the degrees of freedom. Consequently, the 3DGO model is represented as $\mathcal{G}^{obj}=\{\mu_i, r_i, s_i, \alpha_i, h_i\}^{U}_{i=1}$, where $U$ is the number of 3D Gaussians. All segmented reference images with the known poses are utilized to build this 3DGO model. We kindly refer to \cite{kerbl20233d} for more details.

\subsection{Object Pose Inference}
\label{sec_infer}
\begin{figure}[!t]
    \centering
     \includegraphics[width=1.0\linewidth]{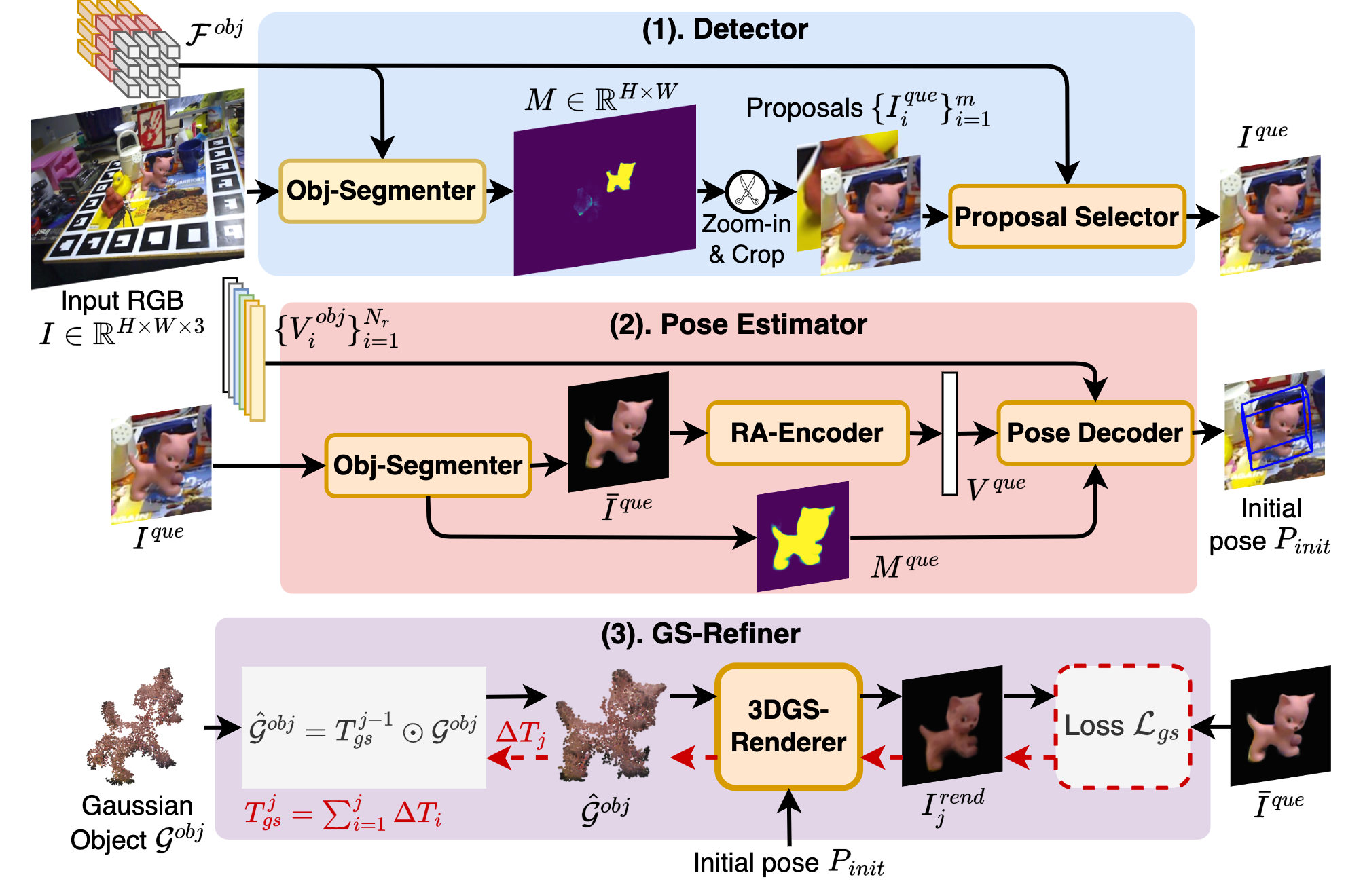}
    \caption{ 
    \textbf{(1). Detector} first employs an Obj-Segmenter to produce a mask from the input image using the semantic information ($\mathcal{F}^{obj}$). Then, connected components are computed from the predicted mask to generate proposals, which are further processed by a proposal selector to determine the final output.
    \textbf{(2). Pose Estimator} utilizes an Obj-Segmenter to predict an object mask $M^{que}$ ($\mathcal{F}^{obj}$ is omitted for clarity). An embedding vector $V^{que}$ is then extracted from the segmented image using RA-Encoder, followed by a pose decoder for estimating an initial pose ($P_{init}$) using both $V^{que}$ and $M^{que}$.
    \textbf{(3). GS-Refiner} starts by applying an optimizable transformation $T^{j-1}_{gs}$ to the 3D coordinates of the 3D Gaussian Object (3DGO) $\mathcal{G}^{obj}$, where $j \geq 1$ is the refinement step.
    Then, the 3D Gaussian Splatting-based renderer (3DGS-Renderer) generates an RGB image ($I_{j}^{rend}$) using the initial pose ($P_{init}$) and the transformed 3DGO ($\hat{\mathcal{G}}^{obj}$). Finally, the gradient $\Delta{T_i}$ is used to update the transformation parameter $T^{j}_{gs}$, minimizing the difference ($\mathcal{L}_{gs}$) between the rendered and the segmented images.
    }
    \label{fig-inference}
\end{figure}

This section outlines the inference pipeline of \gspose, a cascaded process consisting of three core components. 
Firstly, \gspose employs an object detector for detection. Secondly, \gspose obtains an initial pose using a pose estimator based on the detection. Finally, a 3D Gaussian Splatting-based pose refinement module (GS-Refiner) is adopted to optimize the initial pose. \cref{fig-inference} illustrates these components, and we describe each one in detail below.

\boldparagraph{Detector.}
\label{seg-detector}
We leverage a segmentation-based detector to localize the target object (see \cref{fig-inference} top). The detector consists of an Obj-Segmenter (as described in \cref{sec-RA-encoder}) and a proposal selector. Specifically, given an input image, we first apply Obj-Segmenter to predict a segmentation mask, from which we generate a set of mask proposals $\{M^{que}_i\}^{m}_{i=1}$ by finding the connected components, where $m$ represents the number of proposals.
Subsequently, a set of object-centric RGB images $\{I^{que}_i \}^{m}_{i=1}$ are cropped from the input image using the 2D bounding boxes derived from these mask proposals.
Next, we feed these RGB images into the proposal selector to obtain the final detection result.
Within the proposal selector, we first extract the DINOv2 feature tokens $\{F^{que}_i \in \mathbb{R}^{L \times C}\}^m_{i=1}$ from these cropped images and then compute the image-level cosine similarities between these image features and the object semantic representation $\mathcal{F}^{obj}$. We select the one with the highest similarity score as the output, denoted as $I^{que}$.

\boldparagraph{Pose Estimator.} 
In the second stage, we estimate an initial pose using a template retrieval-based pose estimator (see \cref{fig-inference} middle).
This pose estimator is comprised of an Obj-Segmenter (identical to the one in Detector), an RA-Encoder (as described in \cref{sec-RA-encoder}), and a pose decoder. 
We first obtain a segmentation mask $M^{que}$ using Obj-Segmenter as well as an image-level representation vector $V^{que}\in \mathbb{R}^{64}$ using RA-Encoder from the detection.
We then input $M^{que}$ and $V^{que}$ into the pose decoder to compute an initial 6D pose $P_{init}=[R_{init}, t_{init}]$.
More specifically, the pose decoder first computes the cosine similarity scores $\{c_i=||V^{que}|| \cdot ||V^{obj}_i||\}^{N_r}_{i=1}$ between the query vector $V^{que}$ and the reference vectors within the set of 3D object rotation-aware representations $\{V^{obj}_i\}^{N_r}_{i=1}$. Consequently, the reference rotation matrix $R^{ref}_j$ with the highest similarity score is retrieved as the initial 3D rotation estimate ($R_{init}$), where $j$ denotes the index of the closest reference template. We then analytically infer the initial 3D translation $t_{init}$ using the query mask ($M^{que}$) and the $j^{th}$ reference mask ($M^{ref}_j$).
%
Specifically, we calculate a relative scale factor $\delta_s \in \mathbb{R}$ and a relative 2D center offset ratio $\Delta_{xy} \in \mathbb{R}^2$ between $M^{que}$ and $M^{ref}_j$ as follows:
\begin{equation}
\small
\begin{split}
\begin{cases}
        \delta_s    &= \sqrt{{Area}(M^{que})/{Area}(M^{ref}_j)}, \\
        \Delta_{xy} &= ({C_{bbox}}(M^{que}) - {C_{bbox}}(M^{ref}_j)) / S,
\end{cases}
\end{split}
\end{equation}
where ${Area}(M) = \sum\limits_{j=0}^{S}\sum\limits_{i=0}^{S}M_{[i,j]}$ denotes the mask area, \(S\) is the mask scale, and ${C_{bbox}}(M)$ denotes the 2D center of the bounding box tightly surrounding the mask $M$. We then compute the distance $t_z^{que} \in \mathbb{R}$ and the 2D center $P^{que}_{xy}\in \mathbb{R}^2$ of the object in the query image by
\begin{equation}
\small
\begin{split}
     t_z^{que} = t_z^{ref} S / \delta_s / S^{que}_{box}, ~~
     P^{que}_{xy}=S^{que}_{box} \Delta_{xy} + C^{que}_{box},
\end{split}
\end{equation}
where $t_z^{ref}$ is the pre-computed z-axis distance of the object in the reference image (more details provided in the supplementary materials), $C^{que}_{box}$ and $S^{que}_{box}$ are the 2D center and scale of the 2D object bounding box predicted from the original input image. Finally, we obtain the initial translation estimate by 
\[t_{init} = t_z^{que}  K^{-1}\bar{P}^{que}_{xy} ,\]
where $\bar{P}^{que}_{xy} \in \mathbb{R}^3$ is the homogeneous form of $P^{que}_{xy}$, and $K$ denotes the camera intrinsic matrix.

\boldparagraph{GS-Refiner.}
The initial pose estimate is further refined by leveraging the 3D object representation $\mathcal{G}^{obj}$ through an iterative render-and-compare optimization procedure. This pose refinement stage, termed GS-Refiner, utilizes differentiable 3D Gaussian Splatting-based rendering \cite{kerbl20233d}, which facilitates the optimization of a learnable transformation $T_{gs}$ to minimize the discrepancy between the rendered object and the observed query image. Formally, the optimal transformation $T_{gs}^*$ is obtained by minimizing the following objective:
\begin{equation}
\begin{split}
T_{gs}^{*} = \underset{T_{gs}}\argmin \mathcal{L}_{gs}(\mathcal{R}_{gs}(T_{gs} \odot \mathcal{G}^{obj}, P_{init}), \bar{I}^{que}) .
\end{split}
\end{equation}
Here, $\mathcal{R}_{gs}$ denotes the differentiable rendering function, $T_{gs} \in SE(3)$ represents the learnable transformation parameters, $\odot$ indicates applying a rigid transformation to the 3D coordinates of $\mathcal{G}^{obj}$, and $\bar{I}^{que}$ is the segmented query image. The loss function $\mathcal{L}_{gs}$ is defined as a combination of the losses based on the image structural similarity (SSIM) and Multi-Scale SSIM \cite{wang2003multiscale}:
\begin{equation}
\mathcal{L}_{gs} = \mathcal{L}_{D-SSIM} + \mathcal{L}_{D-MSSIM}
\end{equation}
The optimization is initialized with an identity transformation and iteratively updates $T_{gs}^{*}$ using the AdamW optimizer \cite{loshchilov2017decoupled} with the cosine annealing learning rate schedule, starting from $5\times 10^{-3}$ over a maximum of \(N_{gs}\) iterations with 10 warm-up steps. An early-stopping strategy is employed when the refinement loss converges to a predefined threshold \(\eta\). The final refined pose is obtained as $P = P_{init} T_{gs}^{*}$.

\subsection{Training Objective Functions}
\label{sec_loss}
We employ the Binary Cross Entropy (BCE) loss to train both Co-Segmenter ($\mathcal{L}_{coseg}$) and Obj-Segmenter ($\mathcal{L}_{objseg}$) for pixel-wise segmentation prediction, \ie,
\begin{equation}
\small
    \begin{split}
        \mathcal{L}_{coseg} = \mathcal{L}_{BCE}(\mathcal{M}, \bar{\mathcal{M}}),  \text{~~~~}
        \mathcal{L}_{objseg} = \mathcal{L}_{BCE}(M, \bar{M}), \\
    \end{split}
\end{equation}
where $\mathcal{M}$ and $\bar{\mathcal{M}}$ separately denote the predicted and ground truth group-wise segmentation masks, $M$ and $\bar{M}$ are the predicted and ground-truth frame-wise segmentation masks, respectively. Additionally, we adopt the Negative Log-Likelihood (NLL) loss to train RA-Encoder for learning the 3D object rotation-aware representation, defined as:
\begin{equation}
\small
    \begin{split}
        \mathcal{L}_{rot} = -log
        \frac{exp(||V^{que}|| \cdot ||V_p||/{\tau})}{\sum_{j=1}^{N_s} exp(||V^{que}|| \cdot ||V_j||/{\tau})},
    \end{split}
\end{equation} 
where $N_s$ is the number of the reference samples in a batch, $V^{que}$ and $V_j$ are the representation vectors of the query and the $j^{th}$ reference samples, respectively, ${\tau}$ is the temperature, and $p$ is the index of the positive training sample determined by measuring the geodesic distance of the 3D rotation matrices, calculated as:
\begin{equation}
    p=\underset{0\leq j \leq N_s}{\argmin} ~\arccos\frac{trace(R^{que}R^T_j)-1}{2},
\end{equation} 
where $R^{que}$ is the ground truth rotation matrix of the query sample, and $R_j$ is for the $j^{th}$ training sample.
Consequently, the entire network is optimized through a combined loss in an end-to-end manner, 
\begin{equation}
\small
\mathcal{L}_{total} = \lambda_{c} \mathcal{L}_{coseg} + \lambda_{o} \mathcal{L}_{objseg} + \lambda_{r} \mathcal{L}_{rot} ,
\end{equation} 
where $\lambda_{\{c,o,r\}}$ represent the balance weights.

\section{Experiments}
\label{sec:expe}

\noindent{\textbf{Datasets.}}
We utilize the synthetic MegaPose dataset \cite{labbe2022megapose} for training and the real-world datasets LINEMOD \cite{hinterstoisser2012model} and OnePose-LowTexture \cite{he2022onepose++} for evaluation. 
The MegaPose dataset was generated using BlendProc \cite{denninger2019blenderproc} and 1000 diverse objects from the Google Scanned Objects dataset \cite{Downs2022GoogleSO} and includes one million synthetic RGB images. The LINEMOD dataset \cite{hinterstoisser2012model} contains 13 objects and is commonly used for 6D object pose evaluation. 
Following \cite{liu2022gen6d,pan2023learning,sun2022onepose,he2022onepose++}, the training split of LINEMOD is selected as reference data, while the testing split is used for evaluation. OnePose-LowTexture \cite{he2022onepose++} is a challenging dataset with low-texture or texture-less objects, from which eight scanned objects are utilized for evaluation.
Each object was captured by two video sequences with different backgrounds.
We follow OnePose++~\cite{he2022onepose++} and select the first video as the reference and the other as query data.




\boldparagraph{Baseline Methods.} For comparison, we assess \gspose against several state-of-the-art methods: Gen6D \cite{liu2022gen6d}, Cas6D \cite{pan2023learning}, OnePose \cite{sun2022onepose}, OnePose++ \cite{he2022onepose++}, and MFOS \cite{lee2023mfos}.
They take RGB reference images of novel objects with known poses as input to define the object coordinate system and then estimate the 6D pose of these objects from query images without retraining the network parameters.

\begin{table}[b]
\scriptsize
\centering
\setlength{\tabcolsep}{4pt} 
\setlength\extrarowheight{1pt}
\begin{tabular}{l | c | c c c c c | r}
\Xhline{2\arrayrulewidth}
Method & Type & cat & duck & bvise & cam & driller  & Avg. \\
\hline
SRPN-P \cite{li2018high}        & BBox & 11.85 & 1.62  & 18.94 & 2.44  & 8.91  & 8.76  \\ 
SRPN \cite{li2018high}          & BBox & 9.72  & 4.56  & 22.47 & 13.43 & 10.97 & 12.23 \\ 
SRPN-D \cite{li2018high}        & BBox & 22.97 & 1.85  & 49.14 & 17.76 & 18.89 & 22.12 \\ 
OSOP \cite{shugurov2022osop}    & BBox & 32.10 & 34.81 & 26.68 & 24.33 & 21.36 & 27.86 \\ 
Gen6D \cite{liu2022gen6d}       & BBox & 76.99 & 42.15 & 63.33 & 72.92 & 48.78 & 60.83 \\ 
Cas6D \cite{pan2023learning}    & BBox & 79.46 & {67.44} & 66.32 & 76.39 & 59.35 & 69.79 \\
LocPoseNet \cite{zhao2022finer} & BBox & {81.68} & 61.80 & \textbf{79.45} & \textbf{80.50} & 68.31 & 74.35 \\
\hline
{\gspose} (ours)                & Mask & 69.14 & 80.07 & 66.46 & 75.51 & 73.08 & 72.85 \\
\textbf{\gspose} (ours)         & BBox & \textbf{84.44} & \textbf{86.88} & 71.76 & 79.04 & \textbf{80.60} & \textbf{80.54} \\
\Xhline{2\arrayrulewidth}
\end{tabular}
\caption{ 
Quantitative results of the 2D object localization on \textbf{LINEMOD \cite{hinterstoisser2012model}} regarding the mAP@[0.5:0.95](\%) metric. 
"Type" indicates the detection type in the form of either bounding boxes or segmentation masks. \gspose derives the minimum 2D object bounding box from the mask prediction for comparison.
We highlight the best in \textbf{Bold}. 
}
\label{table_LM_mAP}
\end{table}

\begin{table*}[tb]
\small
\begin{center}
\setlength{\tabcolsep}{4pt} 
\begin{tabular}{l | c | c c c c c c c c c c c c c | c}
\Xhline{2\arrayrulewidth}
Method & YOLOv5 & ape & bwise & cam & can & cat & driller & duck & ebox{*} & glue{*} & holep. & iron & lamp & phone & Avg. \\
\hline                                &       & \multicolumn{13}{ c }{{ADD(S)@0.1d}} \\ 
\hline
Gen6D\cite{liu2022gen6d}             &  &   -  & 62.1 & 45.6 &  -   & 40.9 & 48.8 & 16.2 &  -   &  -   &  -   &  -   &  -   &  -   &  -   \\
Gen6D\cite{liu2022gen6d}{\dag}       &  &   -  & 77.0 & 66.7 &  -   & 60.7 & 67.4 & 40.5 & 98.3 & 87.8 &  -   &  -   & 89.8 &  -   &  -   \\
Cas6D\cite{pan2023learning}{\dag}    &  &   -  & 86.3 & 70.1 &  -   & 60.6 & 84.8 & 51.3 & 98.8 & 88.5 &  -   &  -   & 93.4 &  -   &  -   \\
OSOP\cite{shugurov2022osop}          &  & 26.1 & 55.6 & 36.2 & 52.2 & 42.5 & 49.6 & 22.2 & 72.4 & 52.3 & 18.6 & 72.3 & 27.9 & 39.6 & 43.6 \\
OnePose\cite{sun2022onepose}          &\cmark & 11.8 & 92.6 & 88.1 & 77.2 & 47.9 & 74.5 & 34.2 & 71.3 & 37.5 & 54.9 & 89.2 & 87.6 & 60.6 & 63.6 \\
\text{OnePose++}\cite{he2022onepose++}&\cmark & 31.2 & {97.3} & {88.0} & 89.2 & 70.4 & {92.5} & 42.3 & \textbf{99.7} & 48.0 & 69.7 & {97.4} & {97.8} & {76.0} & 76.9 \\
MFOS\cite{lee2023mfos}                &\cmark & 47.2 & 73.5 & 87.5 & 85.7 & 80.2 & {92.4} & 60.8 & {99.6} & 69.7 & \textbf{93.5} & 82.4 & {95.8} & 51.0 & 78.4 \\
PoseMatcher \cite{2023posematcher}    &\cmark & 59.2 & 98.1 & 93.4 & 96.0 & 88.0 & \textbf{98.4} & 54.1 & 97.8 & 91.5 & 73.4 & 97.9 & 98.1 & \textbf{92.1} & 87.5 \\
\hline
\gspose (ours)          &   & 59.6 & 99.6 & 96.0 & 97.6 & \textbf{88.9} & 95.1 & 74.9 & 99.3 & 92.2 & 86.8 & 98.2 & 96.7 & 80.7 & 89.7 \\
\textbf{\gspose} (ours) & \cmark & \textbf{71.0} & \textbf{99.8} & \textbf{98.2} & \textbf{97.7} & 86.7 & {96.2} & \textbf{77.2} & 99.6 & \textbf{98.4} & 87.4 & \textbf{99.2} & \textbf{98.9} & {85.0} & \textbf{92.0} \\
\hline                                &       & \multicolumn{13}{ c }{{Proj@5pix}} \\ 
\hline 
Gen6D \cite{liu2022gen6d}{\dag}        &   &   -  & 82.5 & 90.8 &  -   & 96.1 & 72.4 & 79.7 & 97.8 & 96.2 &  -   &  -   & 91.6 &  -   &  -   \\
Cas6D \cite{pan2023learning}{\dag}     &   &   -  & 93.4 & 96.3 &  -   & \textbf{99.0} & \textbf{95.0} & 93.5 & 98.3 & \textbf{98.8} &  -   &  -   & 96.9 &  -   &  -   \\
OnePose\cite{sun2022onepose}           & \cmark & 35.2 & 94.4 & 96.8 & 87.4 & 77.2 & 76.0 & 73.0 & 89.9 & 55.1 & 79.1 & 92.4 & {88.9} & 69.4 & 78.1 \\
\text{OnePose++}\cite{he2022onepose++} & \cmark & {97.3} & \textbf{99.6} & \textbf{99.6} & \textbf{99.2} & {98.7} & {93.1} & {97.7} & \textbf{98.7} & 51.8 & {98.6}   & {98.9} & \textbf{98.8} & \textbf{94.5} & {94.3} \\
\hline
{\gspose} (ours)          &  & 77.5 & 98.9 & 98.4 & 97.6 & 97.6 & 92.3 & 97.7 & 97.3 & 91.3 & 96.5 & 98.9 & 90.9 & 91.9 & 94.4 \\
\textbf{\gspose} (ours)   & \cmark & \textbf{97.9} & 98.9 & 99.1 & 97.6 & 98.9 & 93.7 & \textbf{97.8} & 97.1 & 97.4 & \textbf{98.8} & \textbf{99.6} & 94.2 & 93.8 & \textbf{97.3} \\
\Xhline{2\arrayrulewidth}
\end{tabular}
\end{center}
\caption{
Quantitative results on \textbf{LINEMOD \cite{hinterstoisser2012model}} regarding the ADD(S)@0.1d and Proj@5pix metrics. \cmark ~indicates using the external YOLOv5 \cite{yolov5} as the object detector. {*} indicates symmetric objects. {\dag} indicates that the method includes a subset of objects of LINEMOD as training data. We highlight the best in \textbf{bold}. "-" indicates unavailable results.
}
\label{table_LM_fullset}
\end{table*}

\begin{table}[tb]
\scriptsize
\begin{center}
\setlength{\tabcolsep}{1.5pt}
\begin{tabular}{l | c | c c c c c c c c | r}
\Xhline{2\arrayrulewidth}
Method  & GTBox & Toy. & Tea. & Cat. & Cam. & Shin. & Molie. & David & Marse. & Avg. \\
\hline
PVNet    \cite{peng2020pvnet}   & & 12.3 & {90.0} & 68.1 & 67.6 & 95.6 & 57.3 & 49.6 & {61.3} & 62.7 \\   
Gen6D    \cite{liu2022gen6d}    &  & 55.5 & 40.0 & 70.0 & 42.2 & 62.7 & 16.6 & 15.8 & 8.1 & 38.9 \\ 
OnePose  \cite{sun2022onepose}  & \cmark & 65.6 & 89.0 & 39.7 & {90.9} & 87.9 & 31.2 & 42.7 & 30.4 & 59.7 \\ 
OnePose++\cite{he2022onepose++} & \cmark & \textbf{89.5} & \textbf{99.1} & {97.2} & \textbf{92.6} & {98.5} & {79.5} & \textbf{97.2} & 57.6 & {88.9} \\
\hline
\gspose\textsubscript{init}      &  & 55.0 & 75.7 & 82.6 & 69.7 & 95.1 & 63.4 & 65.7 & 57.5 & 70.6  \\
\textbf{\gspose} (ours)          &  & 89.3 & 86.7 & \textbf{100.0} & 90.2 & \textbf{99.3} & \textbf{95.9} & 91.7 & \textbf{83.6} & \textbf{92.1} \\
\hline
\Xhline{2\arrayrulewidth}
\end{tabular}
\end{center}
\caption
{
Quantitative results on each object in \textbf{OnePose-LowTexure \cite{he2022onepose++}} regarding the ADD(S)@0.1d metric. 
"\textsubscript{init}" indicates the initial pose estimation results of \gspose. 
"GTBox" indicates the ground truth 2D object bounding boxes. We highlight the best in \textbf{bold}.
}
\label{table_lowtexture}
\end{table}

\boldparagraph{Metrics.} We adopt the widely used ADD \cite{hinterstoisser2012model} metric that measures the average distance between 3D points after being transformed by the ground truth and predicted poses. The ADDS metric is used for symmetric objects, which measures the average distance to the closest point instead of the ground truth point. Following the protocol\cite{hinterstoisser2012model}, we report the average recall rate of ADD(S) within {10\%} of the object diameter, denoted as \textbf{ADD(S)@0.1d}. 
We also compute the 2D projection errors of the points after being transformed by the ground truth and predicted poses. We report the average recall rate within 5 pixels, denoted as \textbf{Proj@5pix}.
In addition, the mean Average Precision (\textbf{mAP})[0.5:0.95](\%) \cite{lin2014microsoft} is reported for evaluating the 2D object localization performance.

\boldparagraph{Configurations.} 
In our experiments, we set the hyperparameters: $N_k=8$, \(N_{gs}=400\), \(L_m=4\), \(\eta=1\times 10^{-4}\), $\tau=0.1$, $\lambda_{c}=1$, $\lambda_{r}=1$, $\lambda_{o}=1$, $N_s=32$, unless otherwise specified. We use the AdamW \cite{loshchilov2017decoupled} solver with the cosine annealing learning rate schedule, starting from $1\times 10^{-4}$ to $1\times 10^{-6}$, to train our framework for 100,000 steps on an Nvidia RTX3090 GPU with batch size 2.

\subsection{Object Detection}

\noindent{\textbf{Experiment Setups.}} Given a set of object-centric RGB images as reference data, the task is to localize the object of interest in query images without fine-tuning the model parameters.


\boldparagraph{Results on LINEMOD.}
We report the quantitative results of the 2D object detection on LINEMOD \cite{hinterstoisser2012model} regarding the mAP@[0.5:0.95](\%) metric in Table \ref{table_LM_mAP}. We primarily compare \gspose against Gen6D \cite{liu2022gen6d}, Cas6D \cite{pan2023learning}, and LocPoseNet \cite{zhao2022finer}, which are the most similar works. Overall, \gspose achieves {72.85\%} mAP and {80.54\%} mAP in terms of 2D segmentation masks and the mask-induced 2D bounding boxes, respectively. Our segmentation-based detection approach outperforms all baseline methods.
It is worth noting that all methods include a subset of held-out objects in LINEMOD as training data and evaluate on the other 5 selected objects, except OSOP~\cite{shugurov2022osop} and \gspose.

\begin{table*}[tb]
\scriptsize
\setlength{\tabcolsep}{2pt}
    \begin{minipage}{0.22\textwidth}
        \centering
        \begin{tabular}{l c }
        \Xhline{2\arrayrulewidth}
         \multirow{2}{*}{Variant} & ADD(S) \\
                                  & @0.1d \\
        \hline
        w/o proposal selector         & 88.95 \\
        w/o $\mathcal{L}_{D-SSIM}$    & 89.44 \\
        w/o $\mathcal{L}_{D-MSSIM}$   & 89.29 \\
        \hline
        \textbf{\gspose} (ours) & \textbf{90.86}  \\
        \Xhline{2\arrayrulewidth}
        \end{tabular}
        \caption{Ablation studies on the \textbf{LINEMOD} subset regarding different variants.}
        \label{table_ablate}
    \end{minipage}
    \hfill
    \begin{minipage}{0.43 \textwidth}
        \centering
        \begin{tabular}{c |c c c c c c }
        \Xhline{2\arrayrulewidth}
        \multirow{2}{*}{Method}
          &  \multicolumn{6}{ c }{Number of reference images ($N_r$)} \\ 
            &  8  &  16  &  32  &  64  & 128  & All ($\sim$180)  \\
        \hline
        Gen6D \cite{liu2022gen6d}       & - & 29.07 & 49.41 & - & - & 62.45 \\
        Cas6D \cite{pan2023learning}    & - & 32.43 & 53.90 & - & - & 70.72  \\
        OnePose++ \cite{he2022onepose++}& - & 31.38 & 54.98 & - & - & 78.10  \\
        \hline
        \textbf{\gspose} (ours) & 49.39 & \textbf{62.50} & \textbf{74.50} & 85.81 & 89.00 & \textbf{90.86} \\
        \Xhline{2\arrayrulewidth}
        \end{tabular}
        \caption{Results on the \textbf{LINEMOD} subset regarding the varying number of reference images in terms of the ADD(S)@0.1d metric.}
        \label{table_refnum}
    \end{minipage}
    \hfill
    \begin{minipage}{0.32\textwidth}
    \centering
    \begin{tabular}{ c | c c c c c}
    \Xhline{2\arrayrulewidth}
    \shortstack{~\\Maximum refin-\\ement steps (\(N_{gs}\))} & \shortstack{\\ 100\\~} & \shortstack{\\200\\~} & \shortstack{\\300\\~} & \shortstack{\\400\\~} & \shortstack{~\\500\\~} \\
         \hline
         ADD(S)@0.1d  & 85.58 & 90.31 & 90.70 & 90.86 & 90.82 \\
         Runtime (ms) & 851 & 936 & 950 & 958 & 966  \\ 
    \Xhline{2\arrayrulewidth}
    \end{tabular}
    \caption{Results on the LINEMOD subset in terms of the ADD(S)@0.1d metric. The refinement process automatically terminates when the loss converges.}
    \label{table_gsrefiner}
    \end{minipage}
\end{table*}

\subsection{Object Pose Estimation} 
\noindent{\textbf{Experiment Setups.}} Given a set of reference RGB images of a novel object with known poses, the task is to estimate the 6D pose of the object in query images without fine-tuning the network parameters. We conduct experiments under two settings: 1) pose estimation without pre-existing 2D bounding boxes and 2) pose estimation within pre-existing 2D bounding boxes. The latter involves estimating the pose from cropped object-centric images acquired either using the YOLOv5 detector \cite{yolov5} (Table \ref{table_LM_fullset}) or by projecting the 3D object bounding boxes using ground truth poses (Table \ref{table_lowtexture}).

\boldparagraph{Results on LINEMOD.}
Table \ref{table_LM_fullset} shows the quantitative results in terms of the ADD(S)@0.1d and Proj@5pix metrics. 
Overall, \gspose achieves impressive 89.7\% ADD(S)@0.1d and {94.4\%} Proj@5pix recalls on average, outperforming all baseline approaches.
When using the 2D detection results predicted by YOLOv5 \cite{yolov5}, as in OnePose \cite{sun2022onepose} and OnePose++ \cite{he2022onepose++}, \gspose further improves the ADD(S)@0.1d metric to 92.0\% and Proj@5pix to 97.3\%, setting new state-of-the-art performance on LINEMOD. 
This advantage is largely attributed to the low-textured or symmetric objects (\eg, \textit{ape, duck, glue}), where the correspondence-based methods like OnePose \cite{sun2022onepose}, OnePose++\cite{he2022onepose++}, MFOS \cite{lee2023mfos}, and PoseMatcher~\cite{2023posematcher} inherently struggle.

\boldparagraph{Results on OnePose-LowTexture.}
We further compare \gspose against the baselines \cite{liu2022gen6d,sun2022onepose,he2022onepose++} on OnePose-LowTexture \cite{he2022onepose++}.
In addition, we also include PVNet \cite{peng2020pvnet}, which trains a single network per object using approximately 5000 rendered images.
Table \ref{table_lowtexture} reports the quantitative results regarding ADD(S)@0.1d and shows new state-of-the-art performance (92.1\%) achieved by \gspose. 
The keypoint-based approach OnePose \cite{sun2022onepose} obtains an average recall of {59.7\%}, which lags behind our initial result (70.6\%) by about 10\% and our refined result (92.1\%) by over 30\%.
OnePose relies on local feature matching to establish the keypoint-based 2D-3D correspondences, making it unreliable for low-textured or texture-less objects in this dataset. 
To alleviate this, OnePose++ \cite{he2022onepose++} employs the keypoint-free LoFTR \cite{sun2021loftr} for feature matching and significantly improves the result to 88.9\%. 
Even though OnePose++ necessitates ground-truth 2D object bounding boxes for evaluation, \gspose still outperforms it using our built-in detector.
Compared to the object-specific pose estimator PVNet \cite{peng2020pvnet}, \gspose outperforms it by a substantial margin.

\subsection{Additional Experiments}
We conduct additional experiments on the LINEMOD subset and report the results in \cref{table_ablate}, \cref{table_refnum}, and \cref{table_gsrefiner}. 

\boldparagraph{Ablation studies.} 
To assess the efficacy of the connected component-based proposal selector in object detection, we remove it from our object detector and then utilize the 2D bounding box derived from the entire segmentation mask as the output. As a result, the ADD(S)@0.1d metric decreases by about 2\%, indicating the proposal selector's efficacy. 
Besides, when either $\mathcal{L}_{D-SSIM}$ or $\mathcal{L}_{D-MSSIM}$ is removed from GS-Refiner, the performance decreases, indicating that both terms contribute positively to the pose refinement. 

\boldparagraph{Number of reference images.}
\gspose consistently achieves better performance with more reference images. When using only 32 reference images, \gspose obtains 74.5\% recall, already comparable to or even outperforming the results achieved by the baseline methods using all reference images.

\boldparagraph{Maximum refinement steps.}
As expected, \gspose achieves consistently better performance with more refinement steps, reaching saturation at up to 400 steps. The refinement process terminates when the loss converges, thus resulting in a nonlinear increase in runtime with more steps.


\boldparagraph{Runtime.} 
\gspose takes about one second to process a single RGB image (with resolution $480 \times 640$) on a desktop with an AMD 835 Ryen 3970X CPU and an Nvidia RTX3090 GPU, in which  $\sim$0.16s for object detection, $\sim$0.01s for pose initialization, and $\sim$0.96s for refinement. 
\gspose employs an iterative, gradient-based optimization process for pose refinement, which improves accuracy but at the cost of computational efficiency. 
In future work, we plan to explore more efficient optimization algorithms, such as the Levenberg-Marquardt algorithm, to accelerate the pose refinement process for \gspose.


\section{Discussion and Conclusion}

This work presents \gspose, an integrated framework for estimating the 6D pose of novel objects in RGB images. \gspose leverages multiple representations of newly added objects to facilitate cascaded sub-tasks: object detection, initial pose estimation, and pose refinement. \gspose is trained once using synthetic RGB images and evaluated on two real-world datasets, LINEMOD and OnePose-LowTexture. The experimental results demonstrate that \gspose achieves state-of-the-art performance on the benchmark datasets and shows promising generalization capabilities to new datasets. However, objects with slender or thin structures may pose challenges for \gspose due to poor segmentation. Future work could be extending \gspose for 6D pose tracking of unseen objects.

\section{Acknowledgement}
This work was supported by the Academy of Finland project \#353139. We also acknowledge CSC - IT Center for Science, Finland, for computational resources.




{
    \small
    \bibliographystyle{ieeenat_fullname}
    \bibliography{main}

\begin{thebibliography}{56}
\providecommand{\natexlab}[1]{#1}
\providecommand{\url}[1]{\texttt{#1}}
\expandafter\ifx\csname urlstyle\endcsname\relax
  \providecommand{\doi}[1]{doi: #1}\else
  \providecommand{\doi}{doi: \begingroup \urlstyle{rm}\Url}\fi

\bibitem[Cai et~al.(2022{\natexlab{a}})Cai, Heikkil{\"a}, and Rahtu]{cai2022ove6d}
Dingding Cai, Janne Heikkil{\"a}, and Esa Rahtu.
\newblock Ove6d: Object viewpoint encoding for depth-based 6d object pose estimation.
\newblock In \emph{Proceedings of the IEEE/CVF Conference on Computer Vision and Pattern Recognition}, pages 6803--6813, 2022{\natexlab{a}}.

\bibitem[Cai et~al.(2022{\natexlab{b}})Cai, Heikkil{\"a}, and Rahtu]{cai2022sc6d}
Dingding Cai, Janne Heikkil{\"a}, and Esa Rahtu.
\newblock Sc6d: Symmetry-agnostic and correspondence-free 6d object pose estimation.
\newblock In \emph{2022 International Conference on 3D Vision (3DV)}, pages 536--546. IEEE, 2022{\natexlab{b}}.

\bibitem[Cai et~al.(2023)Cai, Heikkil{\"a}, and Rahtu]{cai2023msda}
Dingding Cai, Janne Heikkil{\"a}, and Esa Rahtu.
\newblock Msda: Monocular self-supervised domain adaptation for 6d object pose estimation.
\newblock In \emph{Scandinavian Conference on Image Analysis}, pages 467--481. Springer, 2023.

\bibitem[Castro and Kim(2023)]{2023posematcher}
Pedro Castro and Tae-Kyun Kim.
\newblock Posematcher: One-shot 6d object pose estimation by deep feature matching.
\newblock In \emph{Proceedings of the IEEE/CVF International Conference on Computer Vision}, pages 2148--2157, 2023.

\bibitem[Chen et~al.(2020{\natexlab{a}})Chen, Li, Wang, and Xu]{chen2020learning}
Dengsheng Chen, Jun Li, Zheng Wang, and Kai Xu.
\newblock Learning canonical shape space for category-level 6d object pose and size estimation.
\newblock In \emph{Proceedings of the IEEE/CVF conference on computer vision and pattern recognition}, pages 11973--11982, 2020{\natexlab{a}}.

\bibitem[Chen et~al.(2022)Chen, Wang, Wang, Tian, Xiong, and Li]{chen2022epro}
Hansheng Chen, Pichao Wang, Fan Wang, Wei Tian, Lu Xiong, and Hao Li.
\newblock Epro-pnp: Generalized end-to-end probabilistic perspective-n-points for monocular object pose estimation.
\newblock In \emph{Proceedings of the IEEE/CVF Conference on Computer Vision and Pattern Recognition}, pages 2781--2790, 2022.

\bibitem[Chen et~al.(2021)Chen, Jia, Chang, Duan, Shen, and Leonardis]{chen2021fs}
Wei Chen, Xi Jia, Hyung~Jin Chang, Jinming Duan, Linlin Shen, and Ales Leonardis.
\newblock Fs-net: Fast shape-based network for category-level 6d object pose estimation with decoupled rotation mechanism.
\newblock In \emph{Proceedings of the IEEE/CVF Conference on Computer Vision and Pattern Recognition}, pages 1581--1590, 2021.

\bibitem[Chen et~al.(2020{\natexlab{b}})Chen, Dong, Song, Geiger, and Hilliges]{chen2020category}
Xu Chen, Zijian Dong, Jie Song, Andreas Geiger, and Otmar Hilliges.
\newblock Category level object pose estimation via neural analysis-by-synthesis.
\newblock In \emph{European Conference on Computer Vision}, pages 139--156. Springer, 2020{\natexlab{b}}.

\bibitem[Collet et~al.(2011)Collet, Martinez, and Srinivasa]{collet2011moped}
Alvaro Collet, Manuel Martinez, and Siddhartha~S Srinivasa.
\newblock The moped framework: Object recognition and pose estimation for manipulation.
\newblock \emph{The international journal of robotics research}, 30\penalty0 (10):\penalty0 1284--1306, 2011.

\bibitem[Deng et~al.(2020)Deng, Xiang, Mousavian, Eppner, Bretl, and Fox]{deng2020self}
Xinke Deng, Yu Xiang, Arsalan Mousavian, Clemens Eppner, Timothy Bretl, and Dieter Fox.
\newblock Self-supervised 6d object pose estimation for robot manipulation.
\newblock In \emph{2020 IEEE International Conference on Robotics and Automation (ICRA)}, pages 3665--3671. IEEE, 2020.

\bibitem[Denninger et~al.(2019)Denninger, Sundermeyer, Winkelbauer, Zidan, Olefir, Elbadrawy, Lodhi, and Katam]{denninger2019blenderproc}
Maximilian Denninger, Martin Sundermeyer, Dominik Winkelbauer, Youssef Zidan, Dmitry Olefir, Mohamad Elbadrawy, Ahsan Lodhi, and Harinandan Katam.
\newblock Blenderproc.
\newblock \emph{arXiv preprint arXiv:1911.01911}, 2019.

\bibitem[Downs et~al.(2022)Downs, Francis, Koenig, Kinman, Hickman, Reymann, McHugh, and Vanhoucke]{Downs2022GoogleSO}
Laura Downs, Anthony Francis, Nate Koenig, Brandon Kinman, Ryan~Michael Hickman, Krista Reymann, Thomas~Barlow McHugh, and Vincent Vanhoucke.
\newblock Google scanned objects: A high-quality dataset of 3d scanned household items.
\newblock \emph{2022 International Conference on Robotics and Automation (ICRA)}, pages 2553--2560, 2022.

\bibitem[Haugaard and Buch(2021)]{haugaard2021surfemb}
Rasmus~Laurvig Haugaard and Anders~Glent Buch.
\newblock Surfemb: Dense and continuous correspondence distributions for object pose estimation with learnt surface embeddings.
\newblock \emph{arXiv preprint arXiv:2111.13489}, 2021.

\bibitem[He et~al.(2017)He, Gkioxari, Doll{\'a}r, and Girshick]{he2017mask}
Kaiming He, Georgia Gkioxari, Piotr Doll{\'a}r, and Ross Girshick.
\newblock Mask r-cnn.
\newblock In \emph{Proceedings of the IEEE international conference on computer vision}, pages 2961--2969, 2017.

\bibitem[He et~al.(2022)He, Sun, Wang, Huang, Bao, and Zhou]{he2022onepose++}
Xingyi He, Jiaming Sun, Yuang Wang, Di Huang, Hujun Bao, and Xiaowei Zhou.
\newblock Onepose++: Keypoint-free one-shot object pose estimation without cad models.
\newblock \emph{Advances in Neural Information Processing Systems}, 35:\penalty0 35103--35115, 2022.

\bibitem[Hinterstoisser et~al.(2012)Hinterstoisser, Lepetit, Ilic, Holzer, Bradski, Konolige, and Navab]{hinterstoisser2012model}
Stefan Hinterstoisser, Vincent Lepetit, Slobodan Ilic, Stefan Holzer, Gary Bradski, Kurt Konolige, and Nassir Navab.
\newblock Model based training, detection and pose estimation of texture-less 3d objects in heavily cluttered scenes.
\newblock In \emph{Asian conference on computer vision}, pages 548--562. Springer, 2012.

\bibitem[Hodan et~al.(2020)Hodan, Barath, and Matas]{hodan2020epos}
Tomas Hodan, Daniel Barath, and Jiri Matas.
\newblock Epos: Estimating 6d pose of objects with symmetries.
\newblock In \emph{Proceedings of the IEEE/CVF conference on computer vision and pattern recognition}, pages 11703--11712, 2020.

\bibitem[Kerbl et~al.(2023)Kerbl, Kopanas, Leimk{\"u}hler, and Drettakis]{kerbl20233d}
Bernhard Kerbl, Georgios Kopanas, Thomas Leimk{\"u}hler, and George Drettakis.
\newblock 3d gaussian splatting for real-time radiance field rendering.
\newblock \emph{ACM Transactions on Graphics}, 42\penalty0 (4), 2023.

\bibitem[Labb{\'e} et~al.(2020)Labb{\'e}, Carpentier, Aubry, and Sivic]{labbe2020cosypose}
Yann Labb{\'e}, Justin Carpentier, Mathieu Aubry, and Josef Sivic.
\newblock Cosypose: Consistent multi-view multi-object 6d pose estimation.
\newblock In \emph{European Conference on Computer Vision}, pages 574--591. Springer, 2020.

\bibitem[Labb{\'e} et~al.(2022)Labb{\'e}, Manuelli, Mousavian, Tyree, Birchfield, Tremblay, Carpentier, Aubry, Fox, and Sivic]{labbe2022megapose}
Yann Labb{\'e}, Lucas Manuelli, Arsalan Mousavian, Stephen Tyree, Stan Birchfield, Jonathan Tremblay, Justin Carpentier, Mathieu Aubry, Dieter Fox, and Josef Sivic.
\newblock Megapose: 6d pose estimation of novel objects via render \& compare.
\newblock \emph{arXiv preprint arXiv:2212.06870}, 2022.

\bibitem[Lee et~al.(2024)Lee, Cabon, Br{\'e}gier, Yoo, and Revaud]{lee2023mfos}
JongMin Lee, Yohann Cabon, Romain Br{\'e}gier, Sungjoo Yoo, and Jerome Revaud.
\newblock Mfos: Model-free \& one-shot object pose estimation.
\newblock In \emph{Proceedings of the AAAI Conference on Artificial Intelligence}, pages 2911--2919, 2024.

\bibitem[Lepetit et~al.(2009)Lepetit, Moreno-Noguer, and Fua]{lepetit2009epnp}
Vincent Lepetit, Francesc Moreno-Noguer, and Pascal Fua.
\newblock Epnp: An accurate o (n) solution to the pnp problem.
\newblock \emph{International journal of computer vision}, 81\penalty0 (2):\penalty0 155, 2009.

\bibitem[Li et~al.(2018{\natexlab{a}})Li, Yan, Wu, Zhu, and Hu]{li2018high}
Bo Li, Junjie Yan, Wei Wu, Zheng Zhu, and Xiaolin Hu.
\newblock High performance visual tracking with siamese region proposal network.
\newblock In \emph{Proceedings of the IEEE conference on computer vision and pattern recognition}, pages 8971--8980, 2018{\natexlab{a}}.

\bibitem[Li et~al.(2018{\natexlab{b}})Li, Wang, Ji, Xiang, and Fox]{li2018deepim}
Yi Li, Gu Wang, Xiangyang Ji, Yu Xiang, and Dieter Fox.
\newblock Deepim: Deep iterative matching for 6d pose estimation.
\newblock In \emph{Proceedings of the European Conference on Computer Vision (ECCV)}, pages 683--698, 2018{\natexlab{b}}.

\bibitem[Li et~al.(2019)Li, Wang, and Ji]{li2019cdpn}
Zhigang Li, Gu Wang, and Xiangyang Ji.
\newblock Cdpn: Coordinates-based disentangled pose network for real-time rgb-based 6-dof object pose estimation.
\newblock In \emph{Proceedings of the IEEE/CVF International Conference on Computer Vision}, pages 7678--7687, 2019.

\bibitem[Lin et~al.(2023)Lin, Liu, Lu, and Jia]{lin2023sam}
Jiehong Lin, Lihua Liu, Dekun Lu, and Kui Jia.
\newblock Sam-6d: Segment anything model meets zero-shot 6d object pose estimation.
\newblock \emph{arXiv preprint arXiv:2311.15707}, 2023.

\bibitem[Lin et~al.(2014)Lin, Maire, Belongie, Hays, Perona, Ramanan, Doll{\'a}r, and Zitnick]{lin2014microsoft}
Tsung-Yi Lin, Michael Maire, Serge Belongie, James Hays, Pietro Perona, Deva Ramanan, Piotr Doll{\'a}r, and C~Lawrence Zitnick.
\newblock Microsoft coco: Common objects in context.
\newblock In \emph{European conference on computer vision}, pages 740--755. Springer, 2014.

\bibitem[Liu et~al.(2022)Liu, Wen, Peng, Lin, Long, Komura, and Wang]{liu2022gen6d}
Yuan Liu, Yilin Wen, Sida Peng, Cheng Lin, Xiaoxiao Long, Taku Komura, and Wenping Wang.
\newblock Gen6d: Generalizable model-free 6-dof object pose estimation from rgb images.
\newblock In \emph{European Conference on Computer Vision}, pages 298--315. Springer, 2022.

\bibitem[Loshchilov and Hutter(2017)]{loshchilov2017decoupled}
Ilya Loshchilov and Frank Hutter.
\newblock Decoupled weight decay regularization.
\newblock \emph{arXiv preprint arXiv:1711.05101}, 2017.

\bibitem[Marchand et~al.(2015)Marchand, Uchiyama, and Spindler]{marchand2015pose}
Eric Marchand, Hideaki Uchiyama, and Fabien Spindler.
\newblock Pose estimation for augmented reality: a hands-on survey.
\newblock \emph{IEEE transactions on visualization and computer graphics}, 22\penalty0 (12):\penalty0 2633--2651, 2015.

\bibitem[Nguyen et~al.(2022)Nguyen, Hu, Xiao, Salzmann, and Lepetit]{nguyen2022templates}
Van~Nguyen Nguyen, Yinlin Hu, Yang Xiao, Mathieu Salzmann, and Vincent Lepetit.
\newblock Templates for 3d object pose estimation revisited: Generalization to new objects and robustness to occlusions.
\newblock In \emph{Proceedings of the IEEE/CVF conference on computer vision and pattern recognition}, pages 6771--6780, 2022.

\bibitem[Oquab et~al.(2023)Oquab, Darcet, Moutakanni, Vo, Szafraniec, Khalidov, Fernandez, Haziza, Massa, El-Nouby, Howes, Huang, Xu, Sharma, Li, Galuba, Rabbat, Assran, Ballas, Synnaeve, Misra, Jegou, Mairal, Labatut, Joulin, and Bojanowski]{oquab2023dinov2}
Maxime Oquab, Timothée Darcet, Theo Moutakanni, Huy~V. Vo, Marc Szafraniec, Vasil Khalidov, Pierre Fernandez, Daniel Haziza, Francisco Massa, Alaaeldin El-Nouby, Russell Howes, Po-Yao Huang, Hu Xu, Vasu Sharma, Shang-Wen Li, Wojciech Galuba, Mike Rabbat, Mido Assran, Nicolas Ballas, Gabriel Synnaeve, Ishan Misra, Herve Jegou, Julien Mairal, Patrick Labatut, Armand Joulin, and Piotr Bojanowski.
\newblock Dinov2: Learning robust visual features without supervision, 2023.

\bibitem[{\"O}rnek et~al.(2023){\"O}rnek, Labb{\'e}, Tekin, Ma, Keskin, Forster, and Hodan]{ornek2023foundpose}
Evin~P{\i}nar {\"O}rnek, Yann Labb{\'e}, Bugra Tekin, Lingni Ma, Cem Keskin, Christian Forster, and Tomas Hodan.
\newblock Foundpose: Unseen object pose estimation with foundation features.
\newblock \emph{arXiv preprint arXiv:2311.18809}, 2023.

\bibitem[Osokin et~al.(2020)Osokin, Sumin, and Lomakin]{osokin2020os2d}
Anton Osokin, Denis Sumin, and Vasily Lomakin.
\newblock Os2d: One-stage one-shot object detection by matching anchor features.
\newblock In \emph{Computer Vision--ECCV 2020: 16th European Conference, Glasgow, UK, August 23--28, 2020, Proceedings, Part XV 16}, pages 635--652. Springer, 2020.

\bibitem[Pan et~al.(2023)Pan, Fan, Feng, Wang, Li, and Wang]{pan2023learning}
Panwang Pan, Zhiwen Fan, Brandon~Y Feng, Peihao Wang, Chenxin Li, and Zhangyang Wang.
\newblock Learning to estimate 6dof pose from limited data: A few-shot, generalizable approach using rgb images.
\newblock \emph{arXiv preprint arXiv:2306.07598}, 2023.

\bibitem[Park et~al.(2019)Park, Patten, and Vincze]{pix2pose2019}
Kiru Park, Timothy Patten, and Markus Vincze.
\newblock Pix2pose: Pix2pose: Pixel-wise coordinate regression of objects for 6d pose estimation.
\newblock In \emph{The IEEE International Conference on Computer Vision (ICCV)}, 2019.

\bibitem[Park et~al.(2020)Park, Mousavian, Xiang, and Fox]{park2019latentfusion}
Keunhong Park, Arsalan Mousavian, Yu Xiang, and Dieter Fox.
\newblock Latentfusion: End-to-end differentiable reconstruction and rendering for unseen object pose estimation.
\newblock In \emph{Proceedings of the IEEE Conference on Computer Vision and Pattern Recognition}, 2020.

\bibitem[Peng et~al.(2019)Peng, Liu, Huang, Zhou, and Bao]{peng2019pvnet}
Sida Peng, Yuan Liu, Qixing Huang, Xiaowei Zhou, and Hujun Bao.
\newblock Pvnet: Pixel-wise voting network for 6dof pose estimation.
\newblock In \emph{CVPR}, 2019.

\bibitem[Peng et~al.(2020)Peng, Zhou, Liu, Lin, Huang, and Bao]{peng2020pvnet}
Sida Peng, Xiaowei Zhou, Yuan Liu, Haotong Lin, Qixing Huang, and Hujun Bao.
\newblock Pvnet: pixel-wise voting network for 6dof object pose estimation.
\newblock \emph{IEEE Transactions on Pattern Analysis and Machine Intelligence}, 2020.

\bibitem[Qi et~al.(2017)Qi, Yi, Su, and Guibas]{qi2017pointnet++}
Charles~R Qi, Li Yi, Hao Su, and Leonidas~J Guibas.
\newblock Pointnet++: Deep hierarchical feature learning on point sets in a metric space.
\newblock \emph{arXiv preprint arXiv:1706.02413}, 2017.

\bibitem[Redmon et~al.(2016)Redmon, Divvala, Girshick, and Farhadi]{redmon2016you}
Joseph Redmon, Santosh Divvala, Ross Girshick, and Ali Farhadi.
\newblock You only look once: Unified, real-time object detection.
\newblock In \emph{Proceedings of the IEEE conference on computer vision and pattern recognition}, pages 779--788, 2016.

\bibitem[Ren et~al.(2015)Ren, He, Girshick, and Sun]{ren2015faster}
Shaoqing Ren, Kaiming He, Ross Girshick, and Jian Sun.
\newblock Faster r-cnn: Towards real-time object detection with region proposal networks.
\newblock \emph{Advances in neural information processing systems}, 28:\penalty0 91--99, 2015.

\bibitem[Shugurov et~al.(2022)Shugurov, Li, Busam, and Ilic]{shugurov2022osop}
Ivan Shugurov, Fu Li, Benjamin Busam, and Slobodan Ilic.
\newblock Osop: A multi-stage one shot object pose estimation framework.
\newblock In \emph{Proceedings of the IEEE/CVF Conference on Computer Vision and Pattern Recognition}, pages 6835--6844, 2022.

\bibitem[Su et~al.(2019)Su, Rambach, Minaskan, Lesur, Pagani, and Stricker]{su2019deep}
Yongzhi Su, Jason Rambach, Nareg Minaskan, Paul Lesur, Alain Pagani, and Didier Stricker.
\newblock Deep multi-state object pose estimation for augmented reality assembly.
\newblock In \emph{2019 IEEE International Symposium on Mixed and Augmented Reality Adjunct (ISMAR-Adjunct)}, pages 222--227. IEEE, 2019.

\bibitem[Su et~al.(2022)Su, Saleh, Fetzer, Rambach, Navab, Busam, Stricker, and Tombari]{su2022zebrapose}
Yongzhi Su, Mahdi Saleh, Torben Fetzer, Jason Rambach, Nassir Navab, Benjamin Busam, Didier Stricker, and Federico Tombari.
\newblock Zebrapose: Coarse to fine surface encoding for 6dof object pose estimation.
\newblock \emph{arXiv preprint arXiv:2203.09418}, 2022.

\bibitem[Sun et~al.(2021)Sun, Shen, Wang, Bao, and Zhou]{sun2021loftr}
Jiaming Sun, Zehong Shen, Yuang Wang, Hujun Bao, and Xiaowei Zhou.
\newblock Loftr: Detector-free local feature matching with transformers.
\newblock In \emph{Proceedings of the IEEE/CVF conference on computer vision and pattern recognition}, pages 8922--8931, 2021.

\bibitem[Sun et~al.(2022)Sun, Wang, Zhang, He, Zhao, Zhang, and Zhou]{sun2022onepose}
Jiaming Sun, Zihao Wang, Siyu Zhang, Xingyi He, Hongcheng Zhao, Guofeng Zhang, and Xiaowei Zhou.
\newblock Onepose: One-shot object pose estimation without cad models.
\newblock In \emph{Proceedings of the IEEE/CVF Conference on Computer Vision and Pattern Recognition}, pages 6825--6834, 2022.

\bibitem[Sundermeyer et~al.(2018)Sundermeyer, Marton, Durner, Brucker, and Triebel]{aae2018}
Martin Sundermeyer, Zoltan-Csaba Marton, Maximilian Durner, Manuel Brucker, and Rudolph Triebel.
\newblock Implicit 3d orientation learning for 6d object detection from rgb images.
\newblock In \emph{The European Conference on Computer Vision (ECCV)}, 2018.

\bibitem[Ultralytics(2023)]{yolov5}
Ultralytics.
\newblock Yolov5: Real-time object detection, 2023.

\bibitem[Wang et~al.(2021)Wang, Manhardt, Tombari, and Ji]{gdrnet2021}
Gu Wang, Fabian Manhardt, Federico Tombari, and Xiangyang Ji.
\newblock {GDR-Net}: Geometry-guided direct regression network for monocular 6d object pose estimation.
\newblock In \emph{IEEE/CVF Conference on Computer Vision and Pattern Recognition (CVPR)}, pages 16611--16621, 2021.

\bibitem[Wang et~al.(2019)Wang, Sridhar, Huang, Valentin, Song, and Guibas]{wang2019normalized}
He Wang, Srinath Sridhar, Jingwei Huang, Julien Valentin, Shuran Song, and Leonidas~J Guibas.
\newblock Normalized object coordinate space for category-level 6d object pose and size estimation.
\newblock In \emph{Proceedings of the IEEE/CVF Conference on Computer Vision and Pattern Recognition}, pages 2642--2651, 2019.

\bibitem[Wang et~al.(2003)Wang, Simoncelli, and Bovik]{wang2003multiscale}
Zhou Wang, Eero~P Simoncelli, and Alan~C Bovik.
\newblock Multiscale structural similarity for image quality assessment.
\newblock In \emph{The Thrity-Seventh Asilomar Conference on Signals, Systems \& Computers, 2003}, pages 1398--1402. Ieee, 2003.

\bibitem[Xiang et~al.(2018)Xiang, Schmidt, Narayanan, and Fox]{xiang2018posecnn}
Yu Xiang, Tanner Schmidt, Venkatraman Narayanan, and Dieter Fox.
\newblock Posecnn: A convolutional neural network for 6d object pose estimation in cluttered scenes.
\newblock In \emph{Proceedings of Robotics: Science and Systems (RSS)}, 2018.

\bibitem[Xiao et~al.(2019)Xiao, Qiu, Langlois, Aubry, and Marlet]{xiao2019pose}
Yang Xiao, Xuchong Qiu, Pierre-Alain Langlois, Mathieu Aubry, and Renaud Marlet.
\newblock Pose from shape: Deep pose estimation for arbitrary 3d objects.
\newblock \emph{arXiv preprint arXiv:1906.05105}, 2019.

\bibitem[Zhao et~al.(2022{\natexlab{a}})Zhao, Hu, and Salzmann]{zhao2022finer}
Chen Zhao, Yinlin Hu, and Mathieu Salzmann.
\newblock Locposenet: Robust location prior for unseen object pose estimation.
\newblock \emph{arXiv preprint arXiv:2211.16290}, 2022{\natexlab{a}}.

\bibitem[Zhao et~al.(2022{\natexlab{b}})Zhao, Guo, and Lu]{zhao2022semantic}
Yizhou Zhao, Xun Guo, and Yan Lu.
\newblock Semantic-aligned fusion transformer for one-shot object detection.
\newblock In \emph{Proceedings of the IEEE/CVF Conference on Computer Vision and Pattern Recognition}, pages 7601--7611, 2022{\natexlab{b}}.

\end{thebibliography}
}

\clearpage
\setcounter{page}{1}
\maketitlesupplementary


\section{Supplement}

\subsection{Data Capture with ARKit}
We follow OnePose and OnePose++ \cite{sun2022onepose,he2022onepose++} and leverage the off-the-shelf ARKit\footnote{ARKit. https://developer.apple.com/augmented-reality.} to capture the RGB reference images using an iPhone. In our experiments, we take advantage of the ARKit-based OnePoseCap~\cite{sun2022onepose} application initially developed for OnePose \cite{sun2022onepose}. In OnePoseCap, we first manually define a simple \textit{3D bounding box} around a stationary object, which serves as a minimal, 3D CAD model-free geometric proxy. When using this tool to capture the RGB reference sequence of an object, ARKit automatically tracks the frame-wise camera poses with respect to the predefined 3D box based on feature matching. These tracked poses are then transformed into the 3D bounding box coordinate system and used as the 6D object pose annotations. We kindly refer the reader to OnePose/OnePose++ \cite{sun2022onepose,he2022onepose++} for more details.

\subsection{Preprocessing}
\label{sec_preprecess}
In this part, we describe the image pre-processing step for creating the reference database.
Given the reference data  $\{\hat{I}^{ref}_i, R^{ref}_i, t^{ref}_i\}^{N_r}_{i=1}$ of the target object, where $\hat{I}^{ref}_i \in \mathbb{R}^{H\times W \times 3}$ denotes the $i^{th}$ RGB image, 3D rotation matrix $R^{ref}_i$, and 3D translation vector $t^{ref}_i$, we prepocess these reference images to obtain normalized object-centric images $\{{I}^{ref}_i\}^{N_r}_{i=1}$ with a predefined resolution $S\times S$. 

Specifically, we first compute a 2D square bounding box $\{C^{ref}_i, S^{ref}_i\}$ using the ground truth translation vector $t^{ref}_i$ and the camera intrinsic $K^{ref}$, where $C^{ref}_i \in \mathbb{R}^2$ and $S^{ref}_i\in \mathbb{R}$ are the 2D center and scale of the bounding box, respectively. In particular, we generate the 2D box center $C^{ref}_i$ by projecting the 3D translation $t^{ref}_i$ to the image plane using $K^{ref}$ and then compute the scale factor by $S^{ref}_i=d_{obj} \cdot f^{ref} / {tz}^{ref}_i$, where $d_{obj}$ is the diameter of the 3D object bounding box, $f^{ref}$ is the camera focal length, and ${tz}^{ref}_i \in \mathbb{R}$ denotes the z-axis component of the 3D translation vector $t^{ref}_i$. Subsequently, we crop the object region (${I}^{ref}_i$) using the derived bounding box ($\{C^{ref}_i, S^{ref}_i\}$) and rescale it to the fixed size $S$. 
After preprocessing, the object in all normalized reference images is assumed to be located along the camera optical axis $[0, 0, t_z^{ref}]^T$ at the same distance $t^{ref}_z = d_{obj} \cdot f^{ref} / S $. We set $S=224$ in all experiments.



\subsection{Iterative Pose Refinement with GS-Refiner}

We summarize the iterative pose refinement process in Algorithm \ref{alg-GSrefiner}.

\begin{algorithm}[tb]
\small
\begin{algorithmic}[1]
\STATE Input initial pose: \(P^{init}\)
\STATE Input segmented object image: \(\bar{I}^{que}\)
\STATE Initialize the maximum iteration steps: \(N_g=400\)
\STATE Initialize rotation quaternion parameters: \({q}_0=[1, 0, 0, 0]^T \)
\STATE Initialize translation parameters: \({t}_0=[0, 0, 0]^T \)
\STATE Initialize iteration counter: \( i = 0 \)
\STATE Initialize learning rate : \(r_0 = 0.005\)
\STATE Form transformation matrix: \(P_{0} = \textit{MatrixForm}({q}_0, {t}_0)\)
\WHILE{\(i < N_g \)}
    %
    \STATE Transform object: \(\mathcal{G}^{obj}_{P_i} = \textit{RigidTransform} (\mathcal{G}^{obj}, P_i)\)
    \STATE Render object: \( I^{rend}_{P_i} = \textit{GaussianRenderer} (\mathcal{G}^{obj}_{{P_i}}, P^{init})\)
    \STATE Compute loss: \( \mathcal{L}_{gs} = \textit{LossCriterion}(I^{rend}_{P_i}, I^{que}_{obj})\)
    \STATE Compute gradients: \(\delta_{q} = \frac{\partial \mathcal{L}_{gs}}{\partial q_i}, \delta_{t} = \frac{\partial \mathcal{L}_{gs}}{\partial t_i} \)
    \STATE Update LR: \(r_{i+1} = \textit{CosineAnnealingLRScheduler}(r_i)\)
    \STATE Update params: \( {q}_{i+1} = {q}_{i} + r_{i+1} \delta_{q},~ {t}_{i+1} = {t}_{i} + r_{i+1} \delta_{t} \)
    \STATE Transformation matrix: \(P_{i+1} = \textit{MatrixForm}({q}_{i+1}, {t}_{i+1})\)
    \STATE Update iteration counter: \( i = i + 1 \)
    \IF{ \( \mathcal{L}_{gs}\) converges }
    \STATE \textbf{break}
    \ENDIF
\ENDWHILE
\STATE Update initial pose: \(P = P^{init} P_{i}\)
\end{algorithmic}
\caption{Iterative Pose Refinement within GS-Refiner}
\label{alg-GSrefiner}
\end{algorithm}


\subsection{Additional Results on LINEMOD}
Following Gen6D \cite{liu2022gen6d}, we additionally compare \gspose with baselines on a subset of objects in LINEMOD and report the results in Table \ref{table_LM_subset}. \gspose achieves 47.96\% ADD(S)@0.1d without pose refinement and 90.86\% after refinement, surpassing all baseline approaches by a significant margin. It is noteworthy that without using a subset of objects included in LINEMOD (using the same setup as ours) for training, Gen6D achieves 42.72\% accuracy after pose refinement, falling behind even the initial results of \gspose (47.96\%). As an additional experiment, we also leverage the feature volume-based pose refiner (Vol-Refiner) proposed in Gen6D \cite{liu2022gen6d} for pose refinement. Vol-Refiner improves the initial accuracy to 69.71\%, lagging behind 90.86\% achieved with GS-Refiner.

\begin{figure*}[htb]
    \centering
     \includegraphics[width=\linewidth]{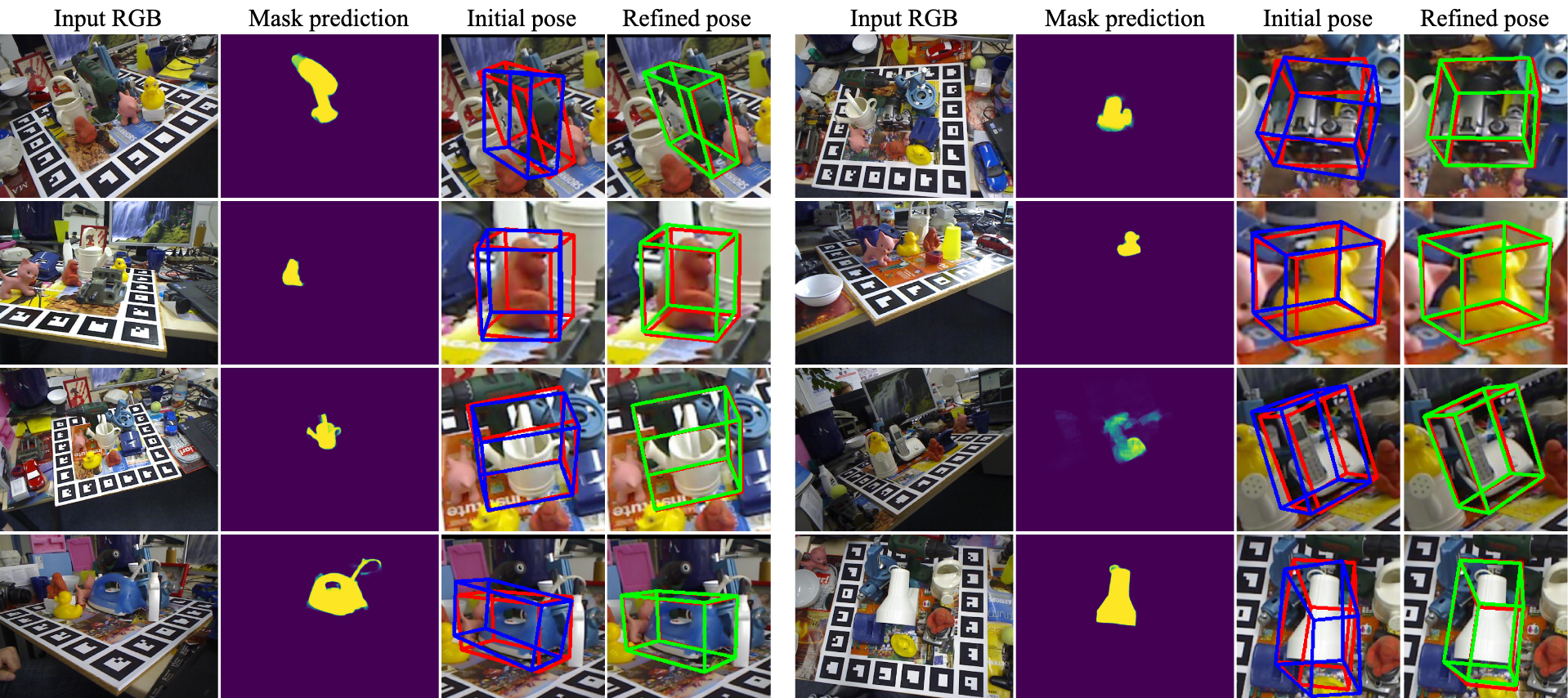}
    \caption{
   Qualitative evaluation on LINEMOD. We present the intermediate segmentation mask predictions (for localization) as well as the estimated 6D poses. Blue, green, and red boxes represent initial, refined, and ground truth poses, respectively. 
    }
    \label{fig_lm_eval}
\end{figure*}

We show qualitative examples from LINEMOD in \cref{fig_lm_eval} and report the complete segmentation and detection results in \cref{table_LM_det_detail} and the initial pose estimation results in \cref{table_LM_detail}.

\begin{table*}[tb]
\begin{center}
\begin{tabular}{l | c | c c c c c | r}
\Xhline{3\arrayrulewidth}
Method & Pose Refinement & cat & duck & bvise & cam & driller & Avg. \\
\hline
Gen6D \cite{liu2022gen6d}{\dag}       & \xmark             & 15.97 & 7.89  & 25.48 & 22.06 & 17.24 & 17.73 \\ 
LocPoseNet \cite{zhao2022finer}{\dag} & \xmark             & - & - & - & - & - & 27.27                     \\
OSOP \cite{shugurov2022osop}          & \xmark             & {34.43} & {20.08} & {50.41} & {32.30} & {43.94} & {36.23} \\
\textbf{\gspose} (ours)              & \xmark              & \textbf{48.70} & \textbf{41.97} & \textbf{55.87} & \textbf{44.31} & \textbf{48.96} & \textbf{47.96} \\
\Xhline{2\arrayrulewidth}
OSOP \cite{shugurov2022osop}           & OSOP \cite{shugurov2022osop}       & 42.54 & 22.16 & 55.59 & 36.21 & 49.57 & 42.21 \\
Gen6D \cite{liu2022gen6d}              & Vol-Refiner \cite{liu2022gen6d}    & 40.92 & 16.24 & 62.11 & 45.59 & 48.76 & 42.72 \\
Gen6D \cite{liu2022gen6d}{\dag}        & Vol-Refiner \cite{liu2022gen6d}    & 60.68 & 40.47 & 77.03 & 66.67 & 67.39 & 62.45 \\
LocPoseNet \cite{zhao2022finer}{\dag}  & Vol-Refiner \cite{liu2022gen6d}    &   -   &   -   &   -   &   -   &   -   & 68.58 \\
Cas6D \cite{pan2023learning}{\dag}     & Cas-Refiner \cite{pan2023learning} & 60.58 & 51.27 & 86.72 & 70.10 & 84.84 & 70.72 \\
\gspose (ours)                       & Vol-Refiner \cite{liu2022gen6d}      & 60.68 & 53.24 & 83.04 & 70.10 & 81.47  & 69.71 \\
\hline
\textbf{\gspose} (ours)              & \textbf{GS-Refiner} (ours)           & \textbf{88.82} & \textbf{74.74} & \textbf{99.61} & \textbf{95.98}  & \textbf{95.14} & \textbf{90.86} \\ 
\Xhline{3\arrayrulewidth}
\end{tabular}
\end{center}
\caption{
Additional quantitative results on the subset of objects in \textbf{LINEMOD \cite{hinterstoisser2012model}} regarding the ADD(S)@0.1d metric. {\dag} indicates that another subset of objects in LINEMOD is included in the training data of the method. "-" indicates unavailable results. We highlight the best in \textbf{Bold}.
}
\label{table_LM_subset}
\end{table*}


\begin{table*}[!h]
\begin{center}
\setlength{\tabcolsep}{3pt} 
\begin{tabular}{l | c | c c c c c c c c c c c c c | r}
\Xhline{3\arrayrulewidth}
Method & YOLOv5 & ape & bwise & cam & can & cat & driller & duck & ebox{*} & glue{*} & holep. & iron & lamp & phone & Avg. \\
\hline
\gspose\textsubscript{init}      &  & 31.5 & 55.9 & 44.3 & 64.9 & 48.7 & 49.0 & 42.0 & 92.8 & 67.7 & 48.1 & 47.2 & 48.9 & 36.0 & 52.1 \\
\textbf{\gspose} (ours)          &  & 59.6 & 99.6 & 96.0 & 97.6 & \textbf{88.9} & 95.1 & 74.9 & 99.3 & 92.2 & 86.8 & 98.2 & 96.7 & 80.7 & 89.7 \\
\hline
\gspose\textsubscript{init}      & \cmark & 39.3 & 58.8 & 45.1 & 64.3 & 53.6 & 50.7 & 38.8 & 93.7 & 74.4 & 52.1 & 55.9 & 56.3 & 37.8 & 55.5 \\
\textbf{\gspose} (ours)          & \cmark & \textbf{71.0} & \textbf{99.8} & \textbf{98.2} & \textbf{97.7} & 86.7 & \textbf{96.2} & \textbf{77.2} & \textbf{99.6} & \textbf{98.4} & \textbf{87.4} & \textbf{99.2} & \textbf{98.9} & \textbf{85.0} & \textbf{92.0} \\
\hline
\Xhline{3\arrayrulewidth}
\end{tabular}
\end{center}
\caption{
Results on \textbf{LINEMOD \cite{hinterstoisser2012model}} regarding the ADD(S)@0.1d metric. 
\cmark ~indicates using the detection results provided by YOLOv5 \cite{yolov5}.
"\textsubscript{init}" indicates the initial pose estimation results of \gspose.
}
\label{table_LM_detail}
\end{table*}

\vspace{1em}

\begin{table*}[!h]
\begin{center}
\setlength{\tabcolsep}{3pt} 
\begin{tabular}{c | c | c c c c c c c c c c c c c | r}
\Xhline{3\arrayrulewidth}
Type & Metric    & ape & bwise & cam & can & cat & driller & duck & ebox & glue & holep. & iron & lamp & phone & Avg. \\
\hline
\multirow{3}{*}{BBox} 
 & mAP@50      & 62.0 & 100.0 & 97.2 & 100.0 & 96.4 & 98.9 & 98.8 & 97.2 & 88.7 & 93.6 & 96.8 & 93.8 & 85.1 & 93.0  \\
 & mAP@75      & 60.8 & 80.6 & 87.8 & 98.8 & 95.1 & 89.5 & 95.1 & 95.7 & 46.1 & 90.0 & 42.0 & 58.4 & 70.9 & 77.8  \\
& mAP@[50:95]  & 56.2 & 71.8 & 79.0 & 87.8 & 84.4 & 80.6 & 86.9 & 81.4 & 53.4 & 74.1 & 52.3 & 56.4 & 61.8 & 71.2 \\
 \Xhline{2\arrayrulewidth}
 \multirow{3}{*}{Mask} 
 & mAP@50      & 66.6 & 100.0 & 98.5 & 100.0 & 96.9 & 99.0 & 99.0 & 98.6 & 89.0 & 95.6 & 100.0 & 94.9 & 91.6 & 94.6 \\ 
 & mAP@75      & 61.9 & 90.9 & 91.6 & 97.3 & 90.6 & 86.9 & 95.1 & 97.0 & 83.5 & 85.9 & 82.3 & 39.6 & 59.0 & 81.7 \\
 & mAP@[50:95] & 53.0 & 66.5 & 75.5 & 70.9 & 69.1 & 73.1 & 80.1 & 79.4 & 61.8 & 67.1 & 62.3 & 47.0 & 53.0 & 71.2 \\
\Xhline{3\arrayrulewidth}
\end{tabular}
\end{center}
\caption{
Complete segmentation and detection results on \textbf{LINEMOD \cite{hinterstoisser2012model}}. "BBox" represents the use of square 2D bounding boxes to evaluate the mask-induced detection results. "Mask" indicates 2D segmentation results.
}
\label{table_LM_det_detail}
\end{table*}





\end{document}